\documentclass{article}
\usepackage[final]{neurips}

\usepackage[utf8]{inputenc} % allow utf-8 input
\usepackage[T1]{fontenc}    % use 8-bit T1 fonts
\usepackage{hyperref}       % hyperlinks
\usepackage{url}            % simple URL typesetting
\usepackage{booktabs}       % professional-quality tables
\usepackage{amsfonts}       % blackboard math symbols
\usepackage{nicefrac}       % compact symbols for 1/2, etc.
\usepackage{microtype}      % microtypography
\usepackage{xcolor}         % colors

\usepackage{makecell}
\usepackage{caption}
\usepackage{subcaption}
\usepackage{tabularx}
\usepackage{svg}
\usepackage{times}
\usepackage{epsfig}
\usepackage{graphicx}
\usepackage{amsmath}
\usepackage{amssymb}
\usepackage{adjustbox}

\usepackage{graphbox}
\usepackage{times}
\usepackage{epsfig}

\usepackage{tikz}
\usetikzlibrary{arrows,shapes}
\usetikzlibrary{shapes.arrows}

\title{Graph Context Encoder: Graph Feature Inpainting for Graph Generation and Self-supervised Pretraining}

\author{%
  Oriel Frigo \\
  AnotherBrain, Paris, France \\
  \texttt{oriel@anotherbrain.ai}
  \And
  Rémy Brossard \\
  AnotherBrain, Paris, France \\
  \texttt{remy@anotherbrain.ai}
  \And
  David Dehaene \\
  AnotherBrain, Paris, France \\
  \texttt{david@anotherbrain.ai}
}

\begin{document}

\maketitle

\begin{abstract}
    We propose the Graph Context Encoder (GCE), a simple but efficient approach for graph representation learning based on graph feature masking and reconstruction.
    
    GCE models are trained to efficiently reconstruct input graphs similarly to a graph autoencoder where node and edge labels are masked. In particular, our model is also allowed to change graph structures by masking and reconstructing graphs augmented by random pseudo-edges.
    
    We show that GCE can be used for novel graph generation, with applications for molecule generation. Used as a pretraining method, we also show that GCE improves baseline performances in supervised classification tasks tested on multiple standard benchmark graph datasets.
    
\end{abstract}

\section{Introduction}

Graphs are useful to model several types of relational data such as molecular structures, protein interactions, social networks and knowledge representation in natural language processing.
Graph Neural Networks (GNNs) recently became popular due to their convincing performance in learning rich representations from graph datasets.  
Inspired by the success of Convolutional Neural Networks (CNNs), the rise of efficient graph convolutions such as the Graph Convolutional Network (GCN) \cite{Kipf2017} and the Graph Isomorphism Network (GIN) \cite{Xu2018} and GINe \cite{Hu*2020Strategies} brought an increasing interest in GNNs with important tasks such as graph generation and graph classification. These tasks can play an increasing role in the field of modern drug discovery, with the example applications of molecule generation task, where one tries to find new drugs given a medication dataset, and the molecular property prediction task, where one tries to predict if a certain molecule is toxic or if it reacts to a certain virus.

Different methods proposed for graph generation could be broadly classified into two main categories: the autoregressive approach and the one shot approach. The autoregressive approach \cite{YouYRHL18, Li2019, Shi*2020GraphAF:} consists in modeling a graph probability distribution by sequential factorization. In practice, reconstruction and generation occurs node by node, edge by edge, potentially requiring several iterations to reconstruct a complete graph. The one-shot approach \cite{Simonovsky2018, DeCao2018, Kwon2019} consists in generating a complete graph in a single neural network pass, as it is typically done in state of the art convolutional generative models on images, such as the GAN \citep{Goodfellow2014} or VAE \citep{Kingma2013VAE}. 

While a considerable number of works focused on autoregressive approaches for graph reconstruction and generation, one shot graph reconstruction received far less attention \cite{Kwon2019}. The main advantage of the one shot approach is its reduced computational complexity, this feature being particularly important in the field of De novo drug design by molecule generation, where the size of datasets can be very large. For instance the ChemBL contains more than 1 million molecules.  However given the challenge of the one shot generation task, the performance of current methods is still quite inferior compared to that of autoregressive methods.   

Indeed, the task of reconstructing a full graph in one shot is not as straightforward as reconstructing grid-like data such as images, as an image can be seen as a graph with fixed connectivity and number of nodes, while general graphs can have arbitrary connectivity and node cardinality.

In this work we explore a different strategy to perform graph generation, by masking and reconstruction. We assume here the task of generating small variations of the training graphs by random perturbations followed by reconstruction. This can be seen as a form of nonparametric sampling, where we make no assumptions about the density probability function of the latent variables explaining the observed graphs. While the method can be used as a one shot procedure, we also consider few shot generation as a midway path between one shot and autoregressive generation.

In addition to graph generation, we also present the self-supervised reconstruction of masked graphs as a proxy task for boosting supervised graph classification. We show that the proposed technique allows an improvement in the performance for different graph neural networks architectures and different datasets.

\section{Related Work}

Applying deep learning methods directly on graph structured data is a recent idea initially proposed by \cite{Gilmer2017}. This initial work mimics the standard convolution layer on images by aggregating the information of local neighbourhood in each node. Although this type of architecture can produce interesting results for tasks such as classification, more complex architectures are required for graph generation, mainly due to the unordered nature of graphs. For example, \cite{Liao2019} uses attention mechanism to generate the graph sequentially. Somehow similarly, \cite{Jin2018} introduced the junction tree approach, which reduces the graph generation to a tree generation problem which is easier to generate, once again sequentially.

This work can be related with the attempts in one shot graph generation \cite{Bresson2019}. In order to deal with the challenging task of single shot generation of graphs with varying number of nodes and edges, \cite{Simonovsky2018} proposes to fix the maximum allowed node cardinality.

The work of \cite{Gao2019} can be seen as an interesting step in the direction of one shot graph reconstruction. Inspired by the U-Net model in computer vision, they propose a model composed of a graph encoder made of graph convolutions and pooling and a graph decoder made of graph convolutions and unpooling. Graph pooling is learned by selecting the top $k$ nodes according to the scalar projection of their features on a learned projection vector. Graph unpooling is performed in a fixed manner by memorizing the structure of the graph encoder at each layer. While Graph Unets are originally applied to node classification task, here we get inspiration from that model to perform node and edge reconstruction.

This paper can also be related to previous works enriching models for graph classification. In \cite{Fey2020} a hierarchical neural message passing architecture is presented for learning on molecular graphs. Two complementary graph representations are provided, a raw molecular graph and an associated junction tree, where nodes representing cycles are clustered together. We show that such model can get benefit from our proposed pretraining.

The recent work of \cite{Hu*2020Strategies} proposes a number of pretraining strategies to improve the performance of graph classification performance, one of these strategies being the node and edge masking. In their method, a number of graph convolutions are performed, followed by a final MLP layer which is used for self-supervised label prediction as a pretraining technique. While their work share some similarities with our proposed method, we propose to perform regression on the full set of graph features and not only masked node label prediction - we perform what is sometimes called blind inpainting where the mask indexes is not given to the model which makes the task potentially harder. Another difference of our work is that we adapt the masking and reconstruction approach for graph generation with the addition of random pseudo-edges for increased variability.

In the computer vision literature, reconstructing masked images is known as the inpainting problem \cite{Bertalmio2000}, in analogy to the task of painting restoration. The work of CNN-based context encoders \cite{pathakCVPR16context} has shown that inpainting can be used as a proxy task for learning rich representations which can be useful to boost the performance of supervised classification. 
While the pretraining strategies for graph neural networks proposed by \cite{Hu*2020Strategies} is mostly inspired by language masking models (e.g. Bert \cite{Devlin2019Bert}), our method can be seen as an adaptation of the CNN-based context encoders approach \cite{pathakCVPR16context} to the domain of graph neural networks.

\section{Proposed method}

We propose a method to perform self-supervised graph reconstruction. The approach consists in randomly masking node and edge features in graphs, and training a graph neural network to predict the ground truth graph features.
One of the main novelties in our approach for graph reconstruction is that we allow the model to reconstruct graph structures which can be different than the input structures, in a single inference shot.

In the following subsections we detail our approach for node-only and node plus edge graph feature masking and prediction. 

\subsection{Node-only feature masking and prediction}
\label{sec:nodemasking}

Node masking consists in corrupting the features of randomly chosen nodes in a given graph. 

Formally, let $\mathcal{G}= (\mathcal{V},\mathcal{E})$ be an undirected graph where $\mathcal{V}$ is a set of nodes and $\mathcal{E} \in \mathcal{V} \times \mathcal{V}$ is a set of edges, $N=|\mathcal{V}|$ is the node cardinality and $M=|\mathcal{E}|$ is the edge cardinality. Graphs are here provided with a set of node label features $ \mathbf{x} = \{\mathbf{x}_1, \mathbf{x}_2, . . . , \mathbf{x}_N\}$ and a set of edge label features $  \mathbf{e} = \{\mathbf{e}_1, \mathbf{e}_2, ..., \mathbf{e}_M \} $.

We take a set of node features $\widetilde{\mathbf{x}}$ initially equal to $\mathbf{x}$ to be masked $K$ times at mask indexes $\Omega$ as follows
\begin{align}
  \mathbf{\widetilde{x}}_{\Omega} = \mathbf{0} ,\, \Omega = \{\omega_1, ..., \omega_K\} \\
  \omega_k \sim \mathcal{U}(1 ,\,N) ,\, k = \{1, ..., K \} 
\end{align}
where $\mathcal{U}$ denotes the uniform distribution. We set to a mask symbol (in practice a vector of zeros) all the features of the masked nodes.

We perform node-only reconstruction through graph convolution, pooling and unpooling similarly to the graph U-nets autoencoder \citep{Gao2019}. We compute $L$ consecutive hidden representations for the masked node features $\widetilde{\mathbf{x}}$ by means of GINe convolutions \cite{Hu*2020Strategies} followed by top $k$ pooling \cite{Gao2019}.  GINe convolutions are computed by
\begin{equation}
\mathbf{\widetilde{x}}^{l*}_i = f_{\mathbf{\Theta}} \left( (1 + \epsilon) \cdot
        \mathbf{\widetilde{x}}^{l}_i + \sum_{j \in \mathcal{N}(i)} \mathrm{ReLU}
        ( \mathbf{\widetilde{x}}^{l}_j + f_\phi(\mathbf{e}_{j,i} )) \right)
\end{equation}
where $\mathcal{N}(i)$ denotes the neighborhood of node $i$ and $f_p$ denote linear MLPs (Multi Layer Perceptrons) parameterized by $p$. First layer node features are initialized as $\mathbf{\widetilde{x}}^{1} = \mathbf{\widetilde{x}}$.
Node pooling is computed as follow:
\begin{align}
\mathbf{y} &= \frac{{\mathbf{\widetilde{x}}^{l*}}\mathbf{p}}{\| \mathbf{p} \|} ,\,
\mathbf{i} = \mathrm{top}_k(\mathbf{y}) \\
\mathbf{\widetilde{x}}^{l+1} &= ({\mathbf{\widetilde{x}}^{l*}} \odot \mathrm{tanh}(\mathbf{y}))_{\mathbf{i}} \\
\mathbf{e}^{l+1} &= \mathbf{e}^{l}_{\mathbf{i},\mathbf{i}}
\end{align}
where $\mathbf{p}$ is a trainable 1D projection vector and $\mathbf{i}$ is the set of pooled node indices. Unpooling is performed as in \cite{Gao2019} by replicating the encoder graph structures into the decoder layers.
We finally minimize a node feature reconstruction loss accounting for the difference between the set of ground truth non-masked features and the set of reconstructed masked features:
\begin{equation}
 \mathcal{L}_n = \sum_{i=1}^N ||\mathbf{x}_i - \widehat{\mathbf{x}_i}||_2 
\end{equation}
where $\mathbf{\widehat{x}} = \mathbf{\widetilde{x}}^{L}$ is the reconstructed set of node features from the last hidden representation layer.  

Note that we can choose to inform the model of which nodes have been masked and predict the feature of only the masked nodes (inpainting task), or leave the model to predict the features of all nodes (blind inpainting task) including the non-masked ones. Here we are interested in the latter, as the reconstruction of the entire graph can be useful for increased variability in graph generation. It can also be seen as a more challenging task in which the model needs more effort in learning useful graph representations. For the best of our knowledge this is the first work proposing blind inpainting on graphs.

\subsection{Incorporating edge feature masking and prediction}
\label{sec:edgemasking}

A major limitation in the usage of graph autoencoder models such as \cite{Gao2019} for one shot graph generation is its inability to predict edge features and to modify graph structures. While being well adapted for node classification tasks, such model is unable to reconstruct an output graph with a different set of edges. 

Here we introduce a novelty to allow the reconstruction of different graph structures given a graph autoencoder architecture. For that, we first augment the graph by adding random pseudo-edges to the input graphs and then we perform message passing that is sensitive not only to node features but also to edge features. 

We perform edge feature masking and prediction in a fashion that allows for a slightly different reconstructed adjacency matrix in comparison to the input adjacency matrix. This can be useful in the task of graph generation, in particular for molecule generation where we search to increase the variety of new molecules while keeping it moderately similar to a set of departing molecules and respecting chemical rules.

In the particular case of molecular graphs, apart from the edge features representing different types of covalent bonds (\emph{single}, \emph{double}, \emph{triple} bonds) we represent \emph{no bond} and \emph{masked bond} as additional edge features. 

%\begin{figure*}
%\centering
%\includegraphics[scale=0.5]{img/schema_masking.png}
%\caption{Illustration of edge and node feature masking. a) Molecular representation highlighting an oxygen atom to be masked. b) Graph representation of the molecule, where black nodes represent carbons, red nodes oxygen and blue node nitrogen. Thin edges represent single bonds, bold edges represent double bonds. c) Addition of random pseudo edges (red edges denote \emph{no bond} edge feature) connected to the node to be masked. c) Node and edge masking, where green edges represents \emph{masked} edge feature and pink node represents \emph{masked} node feature.}
%\label{fig:graphmasking}
%\end{figure*}

\tikzstyle{vertex}=[circle,fill=black,minimum size=5pt,inner sep=0pt]
\tikzstyle{vertex_carbon}=[circle,fill=black,minimum size=5pt,inner sep=0pt]
\tikzstyle{vertex_nitro}=[circle,fill=blue,minimum size=5pt,inner sep=0pt]
\tikzstyle{vertex_ox}=[circle,fill=red,minimum size=5pt,inner sep=0pt]
\tikzstyle{vertex_mask}=[circle,fill=pink,minimum size=5pt,inner sep=0pt]
\tikzstyle{edge} = [draw,-]
\tikzstyle{double_edge} = [draw,thick,-]
\tikzstyle{no_edge} = [draw,-,color=cyan]
\tikzstyle{pseudo_edge} = [draw,-,color=pink]
\tikzstyle{weight} = [font=\small]

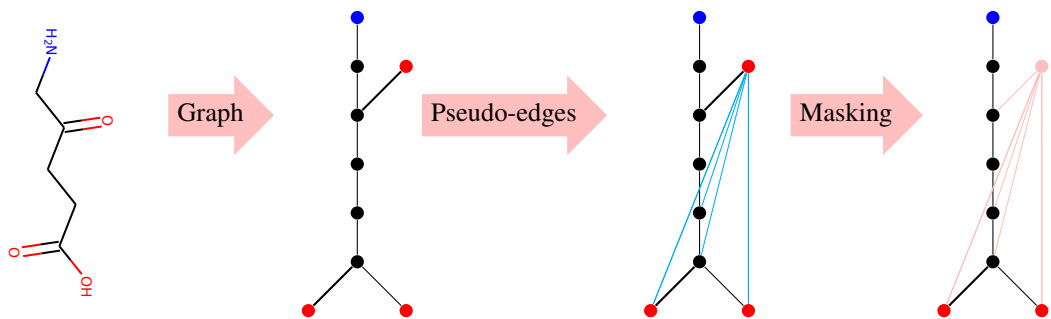
\begin{figure*}
\hspace{1cm}
\begin{tikzpicture}[scale=0.65, auto,swap]

\includegraphics[scale=0.5,vshift=1.5cm]{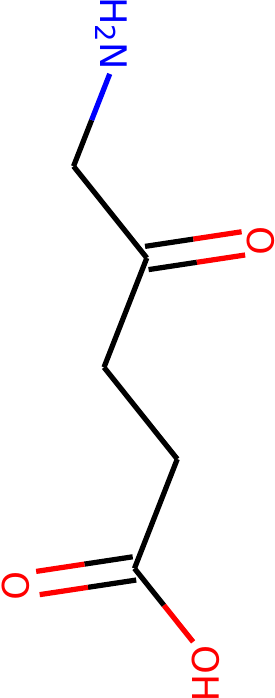}

\node[fill=pink,single arrow] at ({2,6}) {Graph};%
%%% graph 1
   \foreach \pos/\name in {{(5,7)/b}, {(5,6)/c},
                           {(5,5)/e}, {(5,4)/f}, {(5,3)/g}}
       \node[vertex_carbon] (\name) at \pos {};

    \node[vertex_nitro] ({a}) at ({5,8}) {};
    \node[vertex_ox] ({d}) at ({6,7}) {};
    \node[vertex_ox] ({h}) at ({4,2}) {};
    \node[vertex_ox] ({i}) at ({6,2}) {};
       
   \foreach \source/ \dest /\weight in {b/a/1, c/b/1, c/e/1, e/f/1,
                                        f/g/1, g/i/1}
       \path[edge] (\source) -- node[weight] {} (\dest);
\path[double_edge] (g) -- node {} (h);
\path[double_edge] (c) -- node {} (d);

\node[fill=pink,single arrow] at ({8,6}) {Pseudo-edges};%

%\end{tikzpicture}
%\begin{tikzpicture}%
%[every node/.style={single arrow}]\node[fill=cyan] at ({6,7}) {#1};%
%\end{tikzpicture}
%\qquad
%\splarrow{Pseudo-edges} 
%\qquad
%\begin{tikzpicture}
%%% graph 2

   \foreach \pos/\name in {{(12,7)/b}, {(12,6)/c},
                           {(12,5)/e}, {(12,4)/f}, {(12,3)/g}}
       \node[vertex_carbon] ({\name}) at \pos {};

    \node[vertex_nitro] ({a}) at ({12,8}) {};
    \node[vertex_ox] ({d}) at ({13,7}) {};
    \node[vertex_ox] ({h}) at ({11,2}) {};
    \node[vertex_ox] ({i}) at ({13,2}) {};
       
   \foreach \source/ \dest /\weight in {b/a/1, c/b/1, c/e/1, e/f/1,
                                        f/g/1, g/i/1}
       \path[edge] (\source) -- node[weight] {} (\dest);
\path[double_edge] (g) -- node {} (h);
\path[double_edge] (c) -- node {} (d);
\path[no_edge] (d) -- node {} (h);
\path[no_edge] (d) -- node {} (i);
\path[no_edge] (d) -- node {} (h);
\path[no_edge] (d) -- node {} (i);
\path[no_edge] (d) -- node {} (f);
\path[no_edge] (d) -- node {} (g);

%%%

\node[fill=pink,single arrow] at ({15,6}) {Masking};%

%\qquad

%\splarrow{Masking}

%\qquad

d)

%%% graph 3
   \foreach \pos/\name in {{(18,7)/b}, {(18,6)/c},
                           {(18,5)/e}, {(18,4)/f}, {(18,3)/g}}
       \node[vertex_carbon] ({\name}) at \pos {};

    \node[vertex_nitro] ({a}) at ({18,8}) {};
    \node[vertex_mask] ({d}) at ({19,7}) {};
    \node[vertex_ox] ({h}) at ({17,2}) {};
    \node[vertex_ox] ({i}) at ({19,2}) {};
   \foreach \source/ \dest /\weight in {b/a/1, c/b/1, c/e/1, e/f/1,
                                        f/g/1, g/i/1}
       \path[edge] (\source) -- node[weight] {} (\dest);
\path[double_edge] (g) -- node {} (h);
\path[pseudo_edge] (c) -- node {} (d);
\path[pseudo_edge] (d) -- node {} (h);
\path[pseudo_edge] (d) -- node {} (i);
\path[pseudo_edge] (d) -- node {} (h);
\path[pseudo_edge] (d) -- node {} (i);
\path[pseudo_edge] (d) -- node {} (f);
\path[pseudo_edge] (d) -- node {} (g);
\end{tikzpicture}

\caption{Illustration of edge and node feature masking. a) Molecular representation highlighting an oxygen atom to be masked. b) Graph representation of the molecule, where black nodes represent carbons, red nodes oxygen and blue node nitrogen. Thin edges represent single bonds, bold edges represent double bonds. c) Addition of random pseudo edges (cyan edges denote \emph{no bond} edge feature) connected to the node to be masked. d) Node and edge masking, where pink edges represent \emph{masked} edge feature and pink node represents \emph{masked} node feature.}
\label{fig:graphmasking}

\end{figure*}

\subsubsection{Modifying graph structure with random pseudo-edges}

Formally, given a graph with original node features $\mathbf{x}$ and original edge features $\mathbf{e}$, we first perform random masking of node features $\mathbf{x}$ to obtain $\mathbf{\widetilde{x}}$ as explained in the previous section. Then we also obtain a set of masked edge features $\mathbf{\widetilde{e}}$. The simplest and fastest approach to perform edge masking would consist simply in randomly corrupting existent edge features following the same approach of node masking. However, such approach does not allow the reconstructed graphs to increase its degree, in other words, nodes could be disconnected but new connections would not be allowed.
As we are interested in the possibility of moderate changes in the predicted graph structure for the sake of graph generation and for learning features related to graph structure, we propose to connect a number of random “pseudo edges” to masked nodes indexed by $\Omega$ as illustrated in Fig. \ref{fig:graphmasking}. In practice, for each masked node $\mathbf{\widetilde{x}}_{\omega_k}$ where $\omega_k \in \Omega$, a number of $L$ pseudo-edges are connected between $\mathbf{\widetilde{x}}_{\omega_k}$ and a set of randomly chosen nodes. 

Note that with the addition of pseudo-edges, not only the masked graphs may be modified but also the original ground truth graphs. Since we are interested in computing a node-by-node and edge-by-edge feature loss regression (such as in a graph autoencoder), both original and masked graphs should have an adjacency matrix with the same shape. Whenever a pseudo-edge added to the ground truth graph is originally non-existent, it is assigned with a \emph{no bond} edge feature, while the same pseudo-edge added to the masked graph is assigned with a \emph{masked} edge feature.

\subsubsection{Edge feature message passing and loss computation}

In addition to the computation of node hidden representations, we also compute hidden representations for edge features. Edge features from the first layer are first passed though a MLP layer to match the same feature dimensionality of the node features. Edge and node features are then concatenated together.

Inspired by the graph network block presented by \cite{Battaglia2018},  edge feature updates are given by 
\begin{equation}
\mathbf{\widetilde{e}}_{i,j}^{l+1} = f_{\phi} (\mathbf{\widetilde{e}}_{i,j}^{l}, \mathbf{\widetilde{x}}^{l}_i, \mathbf{\widetilde{x}}^{l}_j).
\end{equation}
where $f_{\mathbf{\phi}}$ denote linear MLPs parameterized by $\phi$.

We finally minimize the following total loss composed of node and edge feature reconstruction terms:

\begin{equation}
\mathcal{L} = \frac{1}{N} \sum_{i \in \mathcal{V}} ||\mathbf{x_i} - \mathbf{\widehat{x}_i}||_2 + \lambda \frac{1}{M} \sum_{i,j \in \mathcal{E}} ||\mathbf{e_{i,j}} - \mathbf{\widehat{e}_{i,j}}||_2
\end{equation}

where $\mathbf{\widehat{e}} = \mathbf{\widetilde{e}}^{L}$ is the reconstructed set of edge features and $\lambda$ is a weighting hyperparameter. 

Note that the final reconstructed adjacency matrix is then computed from the reconstructed edge features (e.g. a reconstructed \emph{no bond} edge feature removes an edge from the adjacency matrix), allowing the model to predict different edge connectivity than the input edge connectivity.

\section{Experiments}

In this section, we present a number of experiments performed with our method. In Sec. \ref{sec:molgen} we perform graph reconstruction applied to molecule generation. 

Then, in Sec. \ref{sec:pretrain} we use our model as a pretraining step in order to boost the performance in graph classification tasks. In particular, in Sec. \ref{sec:pretrainmol} we perform molecule reconstruction and applied the pretrained model for molecule property prediction. In Sec. \ref{sec:pretrainmnist} we present experiments on pretraining graph superpixel MNIST dataset to establish the connection with image classification problems.

\subsection{Molecule generation}
\label{sec:molgen}

\begin{figure*}
\centering

\setlength{\tabcolsep}{-1pt}   
\addtolength{\tabcolsep}{-1pt} 

 \begin{tabular}{c || c | c || c | c}
  a) input & b) 1 atom masked & reconstruction & c) 2 atoms masked & reconstruction \\
  \includegraphics[scale=0.3]{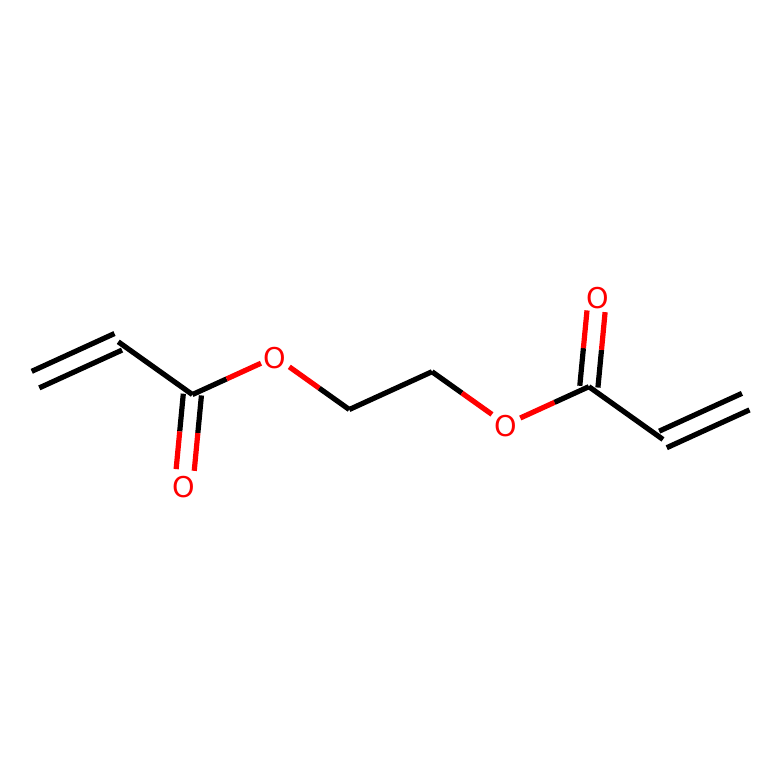} \hspace{0.2cm} & 
  \hspace{0.2cm} \includegraphics[scale=0.3]{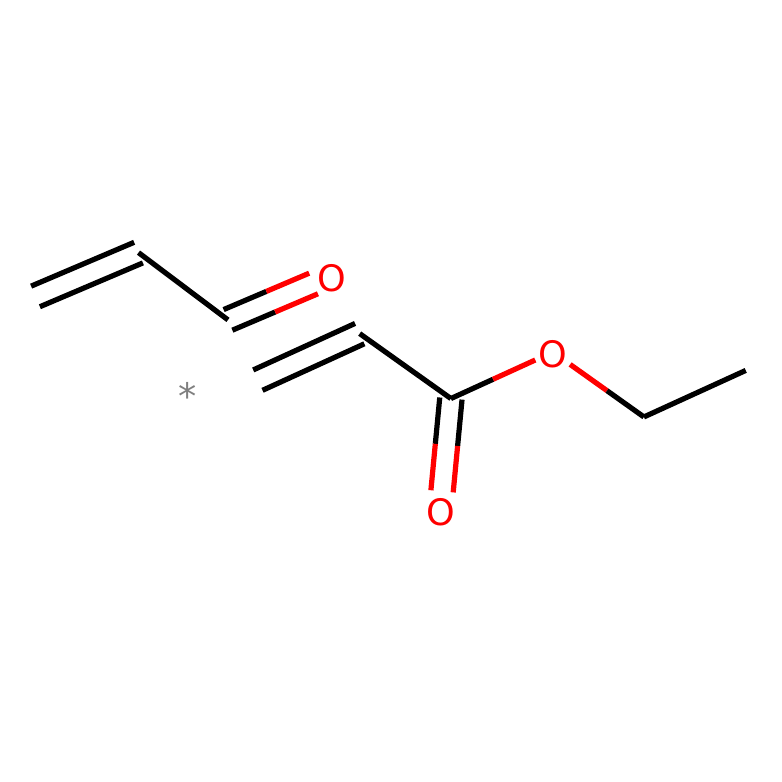} \hspace{0.1cm} & 
  \hspace{0.2cm} \includegraphics[scale=0.3]{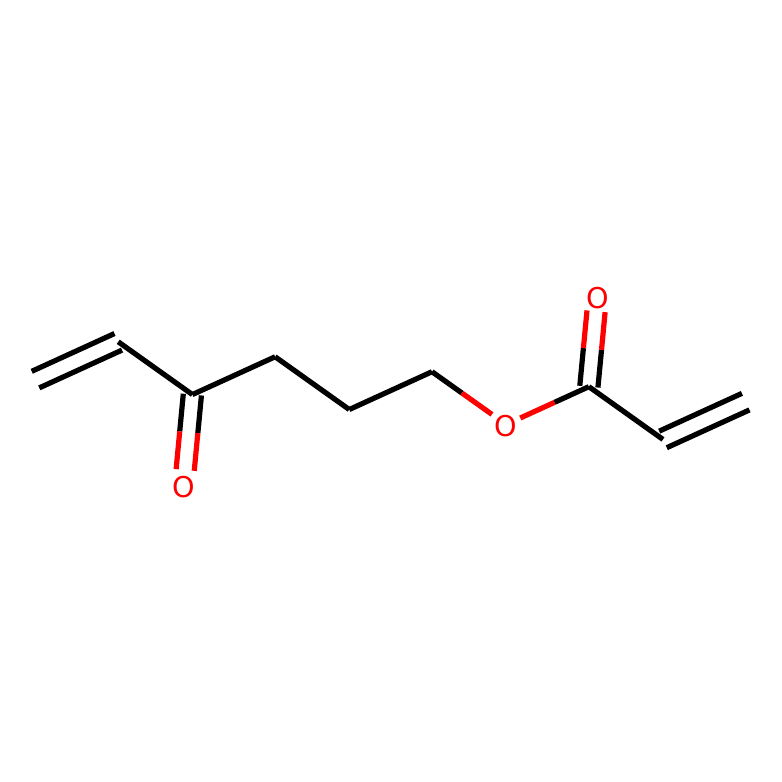} \hspace{0.2cm} &
  \hspace{0.2cm} \includegraphics[scale=0.3]{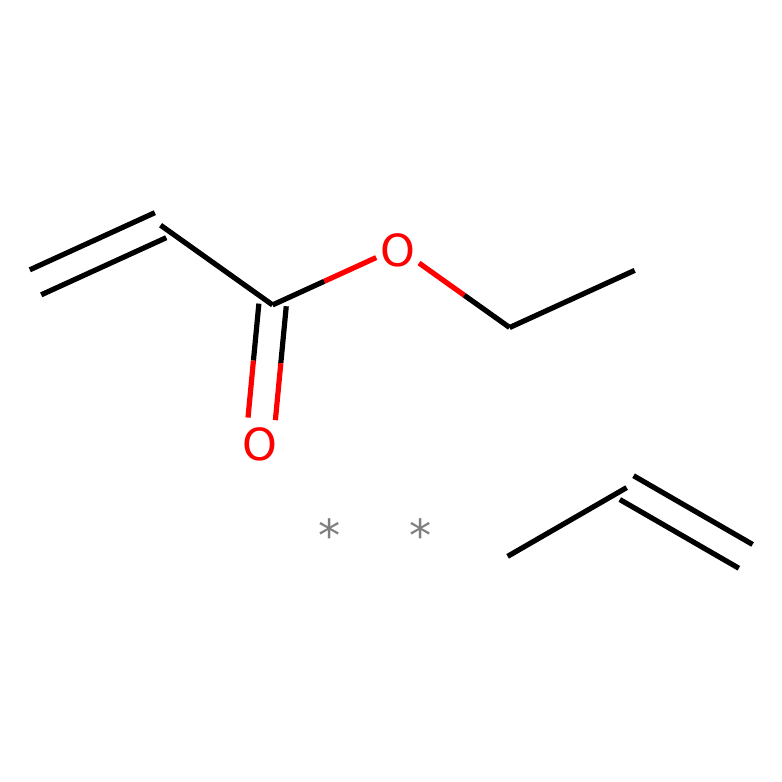} \hspace{0.2cm} &
  \hspace{0.2cm} \includegraphics[scale=0.3]{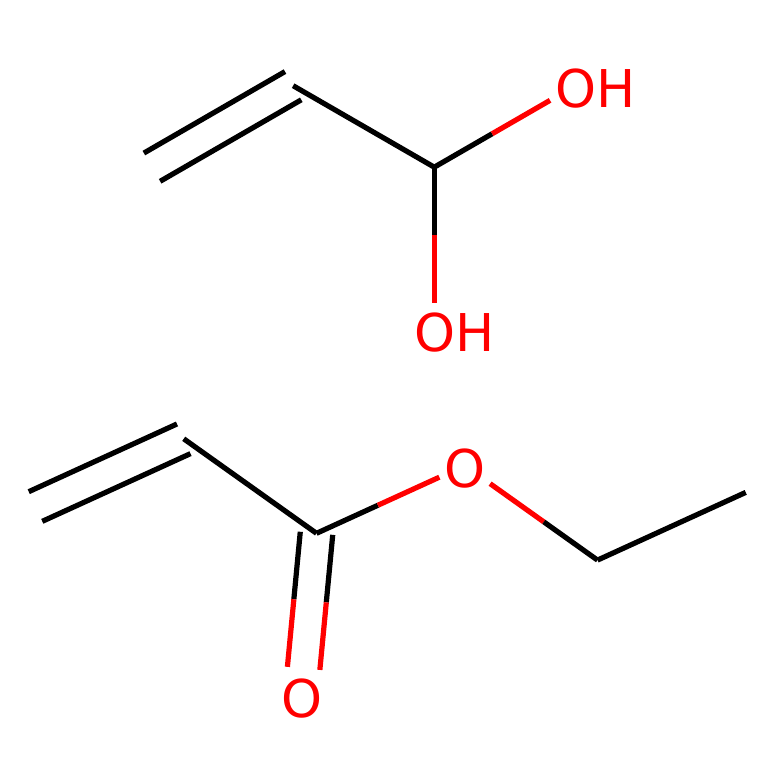} \\
  \hline
  \includegraphics[scale=0.3]{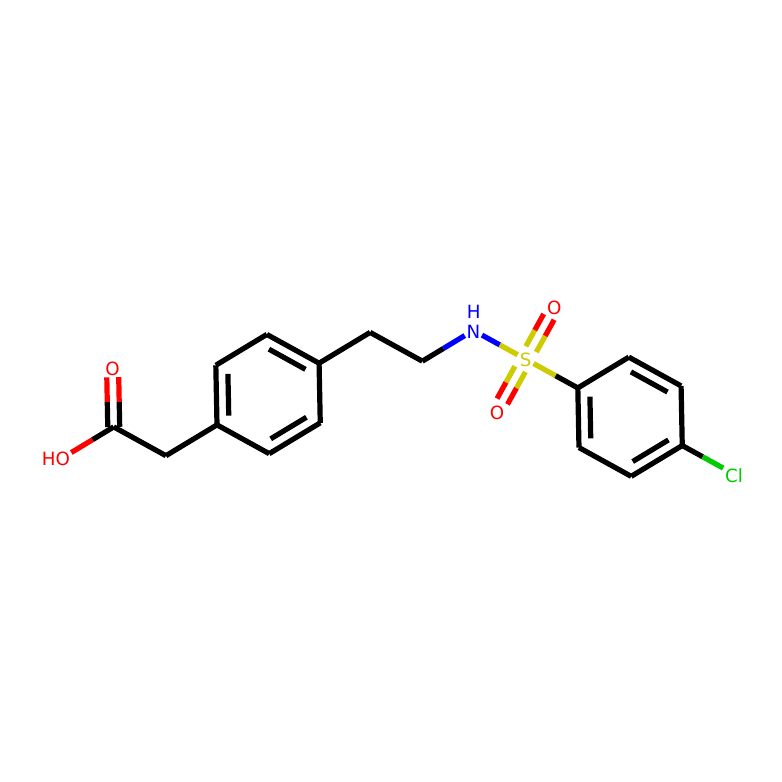} \hspace{0.2cm} & 
  \hspace{0.2cm} \includegraphics[scale=0.3]{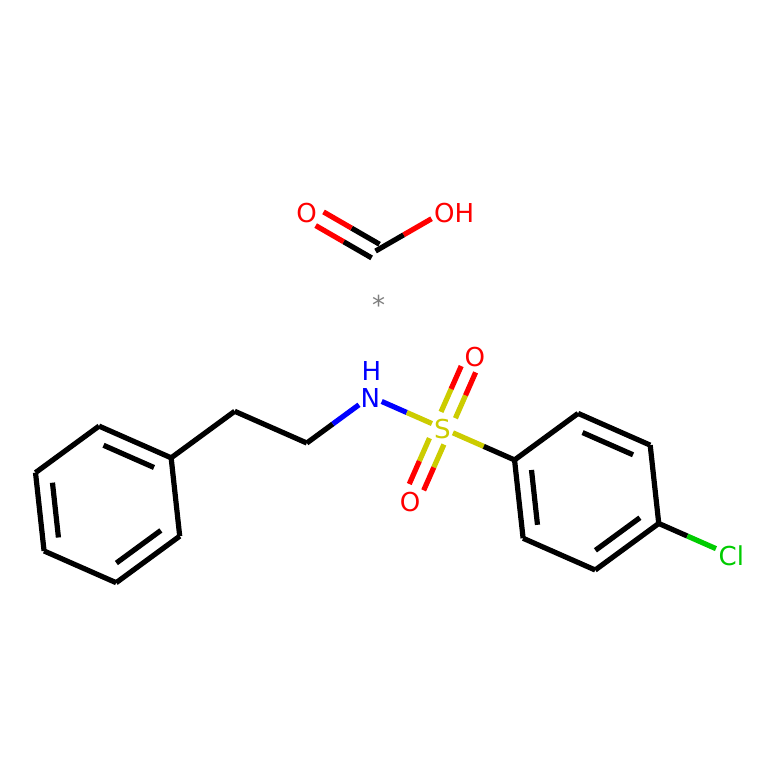} \hspace{0.2cm} & 
  \hspace{0.2cm} \includegraphics[scale=0.3]{img/mol_4.pdf} \hspace{0.2cm} &
  \hspace{0.2cm} \includegraphics[scale=0.3]{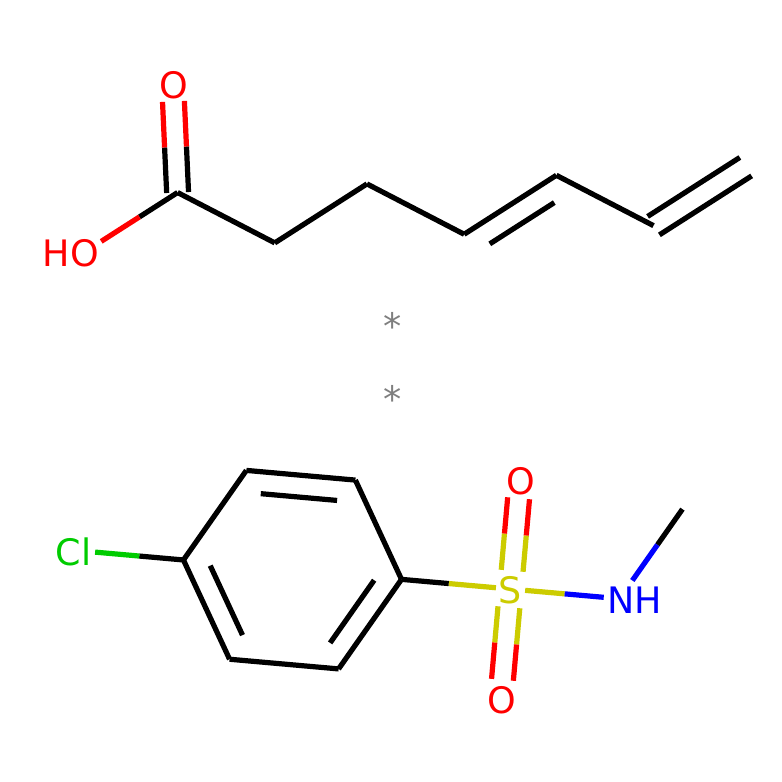} \hspace{0.2cm} &
  \hspace{0.2cm} \includegraphics[scale=0.3]{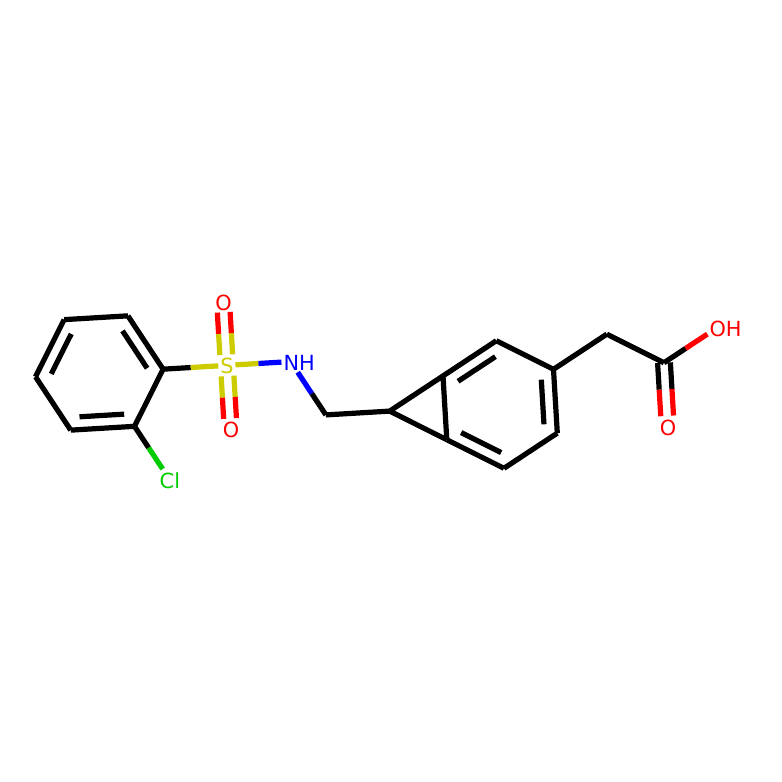} \\
  
\end{tabular}
\caption{Example molecule masking and reconstructions. It can be noted that with only 1 atom masked per molecule, the trained model tends to reconstruct a molecule very similar to the input molecule. On the other hand, when we increase the number of masked atoms to 2, the reconstructed molecule tends to differ more in structure.}
\label{fig:maskingmol} 

\end{figure*}

Differently to generative methods which samples from an approximate graph probability distribution, our method consists in starting with a given set of target molecules and generating variations from these molecules by means of masking nodes and edges and reconstructing it.

We define a one shot generation when a single masking and reconstruction is performed, $n$-shot is defined when the output reconstructed molecule is masked and reconstructed again $n-1$ times, in a recursive manner.

We observed that both the masking rate (amount of masked nodes and edges) and the number of generation shots have a considerable influence on the novelty of generated molecules. In Figure \ref{fig:maskingmol} we show an example of molecule generation with different number of masked nodes. 

Quantitative generation results are summarized in Table \ref{tab:guacamol} where we evaluate our method in a subset of CheMBL dataset with the molecule generation metrics validity, uniqueness, novelty, KL divergence and FCD as proposed by the Guacamol benchmark \cite{Brown2019}. The \emph{validity} assesses whether the generated molecules are actually valid, that is, whether the atom valency is respected. The \emph{uniqueness} benchmark assesses whether models are able to generate non repeated molecules. The \emph{novelty} benchmark penalizes models when they generate molecules already present in the training set. The \emph{KL divergence} measures how well a probability distribution approximates another, where models able to capture the distributions of molecules in the training set will lead to larger KL div. scores. The Frechet ChemNet Distance (FCD) measures how close the generated distributions are to the distribution of molecules in the training set.

For the above mentioned metrics, higher score is better. We compare our performance with the graph based method G-MCTS and the NLP-based methods SMILES LSTSM, AAE, ORGAN and SMILES VAE (values reported by \cite{Brown2019}).

It can be noted in Table \ref{tab:guacamol} that the main strength of the GCE method for molecule generation is its high similarity with the target molecules (high KL divergence score), being competitive with NLP based approaches and outperforming in this score the G-MCTS which is also graph-based like our method, with the disadvantage of performing costly Monte Carlo tree searches for generation. Our novelty ratio score is less competitive but it can be increased in function of the number of inference shots (GCE-2, GCE-3) without affecting much the KL div score. 

When compared to GraphVAE \citep{Simonovsky2018}, a previous graph-based one-shot approach, our method achieves higher validity rates (0.9 in average against 0.3) and is not limited to generate only small graphs. Note that other generation metrics have not been reported for GraphVAE.

In general, our approach remains competitive with previous molecule generation models, with the advantage of a fast generation. While autoregressive or NLP based models would typically need to run at least $M$+$N$ (number of edges + number of nodes) inference shots to generate a single molecule, we show that satisfactory generation can be achieved with only 1, 2 or 3 shots over a large batch of graphs with the proposed approach.

To give a rough figure of the important speed gain when performing molecule generation with our proposed approach, given a trained model, GCE-1 takes less than 10 minutes\footnote{Running time computed on a Nvidia RTX 2080 TI with 8 CPU cores.} to generate 10000 molecules, while JT-VAE \citep{Jin2018} takes more than a dozen hours and the Graph autoregressive flow \citep{Shi*2020GraphAF:} takes a couple of hours to perform the same task, according to the authors.\footnote{The authors report respectively 24 hours and 4 hours on a Tesla V100 GPU with 32 CPU cores. From our experiments with the code provided by \cite{Shi*2020GraphAF:} on a Nvidia RTX 2080 TI, GraphAF took approximately 2 hours to generate 10k molecules.}. 

\begin{table*}
\centering
\begin{tabular}{ c | c | c | c | c | c || c | c | c }
& S-LSTM &	G-MCTS & AAE	& ORGAN	& S-VAE & GCE-1 & GCE-2 & GCE-3 \\
\hline
validity &	0.95 &	1.00 &	0.82 &	0.37 &	0.87 & 0.93 & 0.90 & 0.88 \\
uniqueness &	1.00 &	1.00 &	1.00 &	0.84 &	0.99 & 0.93 & 0.90 & 0.88 \\
novelty  &	0.91 &	0.99 &	0.99 &	0.68 &	0.97 & 0.40 & 0.54 & 0.58 \\
KL div. score  &	0.99 &	0.52 &	0.88 &	0.26 &	0.98 & 0.98 & 0.99 & 0.98\\
FCD score	 &	0.913 &	0.015 &	0.529 &	0.000 &	0.863 & 0.74 & 0.76 & 0.65 \\

\hline
\end{tabular}
\caption{Molecule generation evaluation using a number of metrics proposed by the Guacamol benchmark on a subset of ChemBL dataset. Higher scores are better. GCE-1, GCE-2, GCE-3 denotes our proposed method with respectively 1, 2 and 3 reconstruction shots}

\label{tab:guacamol}

\end{table*}

\subsection{Pretraining for graph classification}
\label{sec:pretrain}

In this section, we show that graph feature inpainting can be applied to pre-train graph representations for a downstream task such as graph classification.

The pretraining technique consists simply in first training a GCE node masking and reconstruction model as described in Section \ref{sec:nodemasking}, and then reusing the model weights to initialize a supervised classifier. The classifier consists of the same GCE architecture with the addition of a final layer consisting of a global mean pooling and a MLP for label prediction. 

We perform experiments with two very different categories of graph datasets, the first one being the MoleculeNet set of datasets containing molecular graphs and the second one being the Graph MNIST dataset containing graphs representing visual digits. 

\subsubsection{MoleculeNet datasets}
\label{sec:pretrainmol}

We perform experiments with the MoleculeNet datasets using the evaluation metrics and datasets provided by the Open Graph Benchmark (OGB) \cite{hu2020open}. 

For this task, we experiment with graph neural networks architectures which apply no pooling function, the GINe \cite{Hu*2020Strategies} and HIMP (Hierarchical Inter-Message Passing network) \cite{Fey2020} models by performing graph context encoding pre-training before performing graph classification. 

For a fair comparison with HIMP, we run the code provided by the authors with the original setup and parameters for each dataset, without any particular grid search for hyperparameter optimization (which may explain the differences in performance compared to the values reported by the authors). For HIMP-GCE, we take the original authors implementation and we load the model weights from a GCE pretrained model with the same architecture.

In Table \ref{tab:moleculenet} we present the results for the datasets MUV, Tox21, ToxCast, HIV and PCBA. All of these datasets are split with 80\% for training and 20\% for test and validation samples. Following \cite{Fey2020}, random splits are performed for MUV, Tox21, ToxCast and HIV and scaffold split is performed for PCBA dataset.
We can see that most baseline models benefit from an increase in performance by using our proposed pretraining technique. It is interesting to note that the GINe model seems to mostly benefit from the pretraining, where we see an increase in performance in 4 out of 5 of the tested datasets. 

\begin{table*}
\centering
\begin{tabular}{ c | c | c || c | c}
%\begin{tabular}{ cccccc}
& GINe & GINe-GCE (our) & HIMP & HIMP-GCE (our) \\ 
\hline
 MUV & $\textbf{0.8599} \pm 0.0109$ & $0.8456 \pm 0.0097$ & $\textbf{0.8043} \pm 0.0228$ & $0.7956 \pm 0.0076$ \\ % $0.7764 \pm 0.0292$ \\ 
Tox21 & $0.8323 \pm 0.0035$ &  $\textbf{0.8450} \pm 0.0019$ &  $0.8437 \pm 0.0068$ & $\textbf{0.8571} \pm 0.0046$ \\ 
ToxCast & $0.6908 \pm 0.0073$  & $\textbf{0.6950} \pm 0.0015$  & $\textbf{0.7218} \pm 0.0048$ & $0.7179 \pm 0.0059$ \\ 
HIV & $0.7999 \pm 0.0161$ & $\textbf{0.8034} \pm 0.0168$ & $0.8109 \pm 0.0090$ & $\textbf{0.8318} \pm 0.0060$ \\ 
PCBA &  $0.2661 \pm 0.0030$ & $\textbf{0.2689} \pm 0.0036$ & $0.2740 \pm 0.002$ & $\textbf{0.2774} \pm 0.0012$ \\
\hline
\end{tabular}
\caption{\label{tab:moleculenet} MoleculeNet results for GINe and HIMP models augmented by our proposed GCE pretraining. Following the setup of \cite{Fey2020}, for dataset PCBA we present average precision results, for all remaining datasets we present the AUROC results. In all cases the higher the score, the better. It can be seen that our proposed GCE pretraining tends to improve the classification performance of the two different baselines.}
\end{table*}

%\emph{to do}

\subsubsection{Graph MNIST dataset}
\label{sec:pretrainmnist}

In this section, we describe the experiments performed with the Graph MNIST dataset made available by \cite{Monti2017}. We show that graph context encoder pretraining can be also useful for vision applications.
The dataset is obtained from the well known MNIST dataset \cite{lecun2010mnist} containing 60000 training and 10,000 test samples. Graphs are computed from a superpixel clustering of images, where each graph contains 75 nodes representing cluster centroids and edges represent the cluster intersections. 

Using the node-only approach described in Sec. \ref{sec:edgemasking}, we perform node masking by corrupting the node features (composed of gray level intensity and spatial position). Edge feature masking is not performed for this dataset, since the dataset is not originally provided with edge features. 

Results presented in Table \ref{tab:graphmnist}, hints a considerable improvement in performance when a GCE pretrained model is applied for digit classification.

\begin{figure*}
\centering
\includegraphics[scale=1.5]{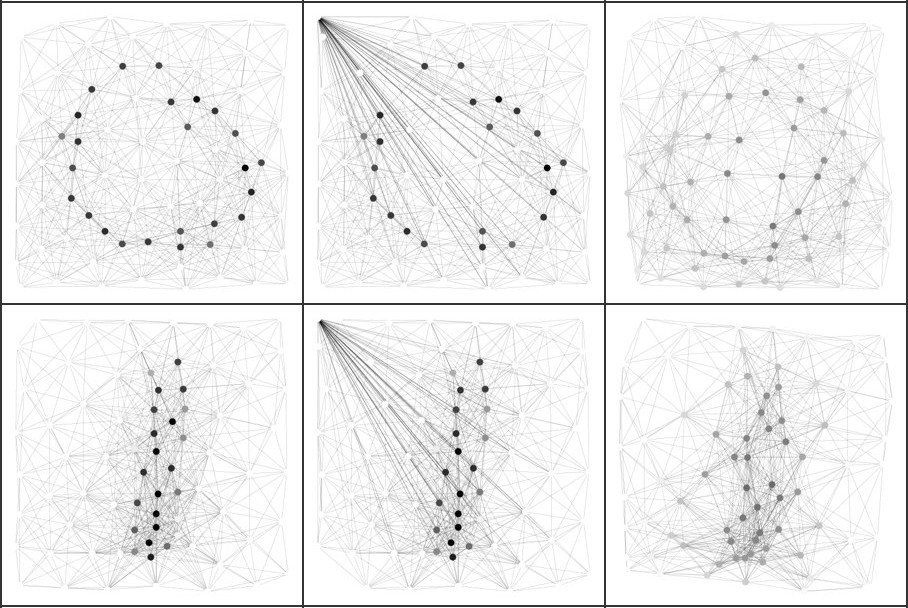}
\caption{Masking and reconstruction of digits on Graph MNIST test dataset. Left column: original graphs representing digits 0 and 1; center column: graphs with a number of nodes masked; right column: reconstructed graphs. It can be noted that the reconstruction resembles the original graphs but is far from perfect, as the model is not informed of which nodes have been masked. Nevertheless, the learned graph representations can be used to improve digit classification performance, as shown in Table \ref{tab:graphmnist}. }
\end{figure*}

\begin{table}
\centering
\begin{tabular}{ c  c | c }
& Method & Classification Error (\%) \\ 
\hline
%Graph MNIST w/out pos. &  &  &  &  \\ 
& Monet & $8.89$ \\
& Spline CNN & $4.78$  \\ 
& GUnet & $3.53 \pm 0.11 $  \\ 
& GUnet-GCE & $\textbf{1.90} \pm 0.11 $  \\ 

\hline
\end{tabular}
\caption{\label{tab:graphmnist} Graph MNIST classification comparison with baselines Monet \cite{Monti2017}, Spline CNN \cite{Fey2018}, Graph UNet \cite{Gao2019} and Graph UNet augmented by our proposed GCE pretraining. It can be noted that our pretraining strategy allows for a considerable improvement in the performance of digit classification compared to other baselines.}
\end{table}

\subsection{Implementation and setup details}

The algorithm described in this paper was implemented using pytorch \citep{NEURIPS2019_9015} and torch geometric libraries \citep{Fey2019torchgeo}. For molecule generation we set $L := 6$ (3 encoder and 3 decoder layers) with $50$ hidden channels. 

For molecule generation we use Graph Unet architecture, with top $k$ pooling set to a pooling rate of $0.5$ between layers, and residual connections between the encoder and decoder. For GINe we set the pooling rate to $1$ so that graphs are not downsampled. We apply a masking rate of 10\% for node and edge features.

For molecule property prediction in MoleculeNet datasets, we use the setup parameters (dataset splits, metrics and network architecture hyperparameters) provided by the authors official implementation of HIMP \cite{Fey2020}, available online. 

Whenever we deal with molecule datasets, each graph represents a molecule, where nodes are atoms, and edges are chemical bonds. Input node features contain atomic number and chirality, as well as other additional atom features such as formal charge and whether an atom is in a ring. Edge features contain the bond information (single, double, triple, masked, no bond). Node and edge features are represented by one-hot encoding.

For experiments with Graph MNIST, we use the standard train, test, and validation partition from the original dataset. We set $L := 4$ (2 encoder and 2 decoder layers) with $50$ hidden channels.

For optimization, we used the adaptive stochastic gradient descent ADAM algorithm \cite{kingma2014adam} with a learning rate set to $10^{-2}$ and we train for 100 epochs. For inpainting masked graphs we set the edge feature weight $\lambda := 2$ in the reconstruction loss. For supervised classification experiments, we minimize the cross entropy loss. Whenever we report classification results with mean $\pm$ standard deviation, we run 5 times the training and evaluation steps.

\section{Conclusion}

In this work, we have proposed a method for self-supervised graph reconstruction based on graph feature masking.

We have explored two main applications for this method. First, as a nonparametric generation  method by masking and reconstruction. Notably, our method introduces novel graph pseudo-edge augmentations allowing a graph autoencoder to reconstruct graphs with a different structure than its input in a computationally efficient manner. 

We consider that our proposed graph generation approach may be particularly appealing in applications where we search for automated variations of a number of target graphs. As our method can be tuned in favor of more or less novelty, the desired degree of similarity with the target graphs can be easily tuned in function of the masking rate and the number of reconstruction shots.

Note that our main interest here is to perform very fast and efficient sampling of moderate variations of a set of target molecules, in a single or in a few inference shots, even at the cost of degrading some of the molecule generation metrics.

While the proposed molecule generation method is limited in the sense of not being currently adapted for molecular optimization, future work could be employed in reconstructing molecules constrained to increase the score of molecular properties such as druglikeness.

Second, we have shown that our method can be useful as a pretraining proxy task for boosting the performance of graph classification. This is confirmed with several experiments on different datasets: the graph molecular datasets MoleculeNet and in the vision dataset graph MNIST.

From our experiments we observed that graph pooling and unpooling were not always beneficial in improving graph property prediction due to the fact that the employed unpooling method is not learned. Hence, the combination of our method with differentiable pooling/unpooling methods could be a promising road for future work.

\bibliographystyle{unsrtnat}
\bibliography{graph_context_encoder}

\appendix

\section{Sample generated molecules}

\begin{figure}[h]
\begin{tabular}{cccccc}
  \includegraphics[scale=0.18]{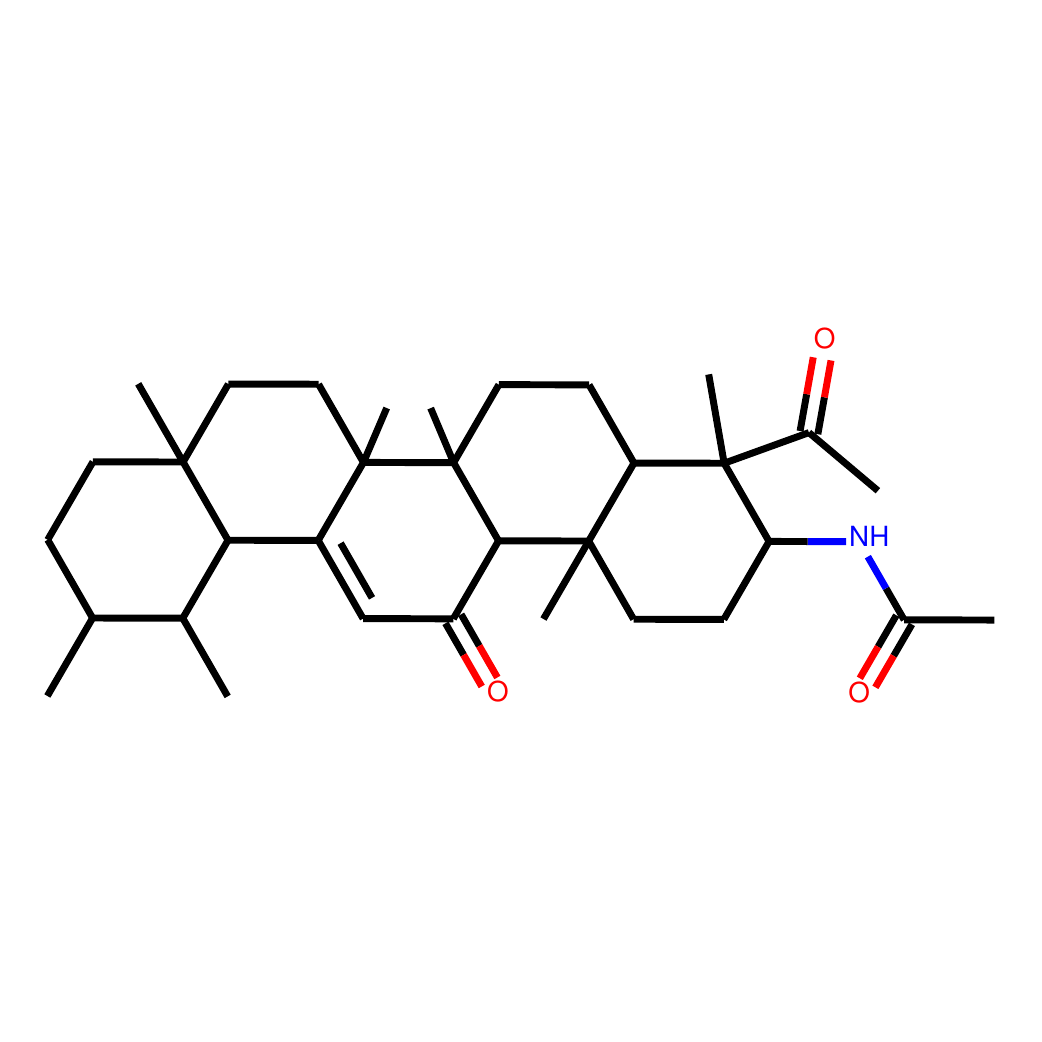} &
  \includegraphics[scale=0.18]{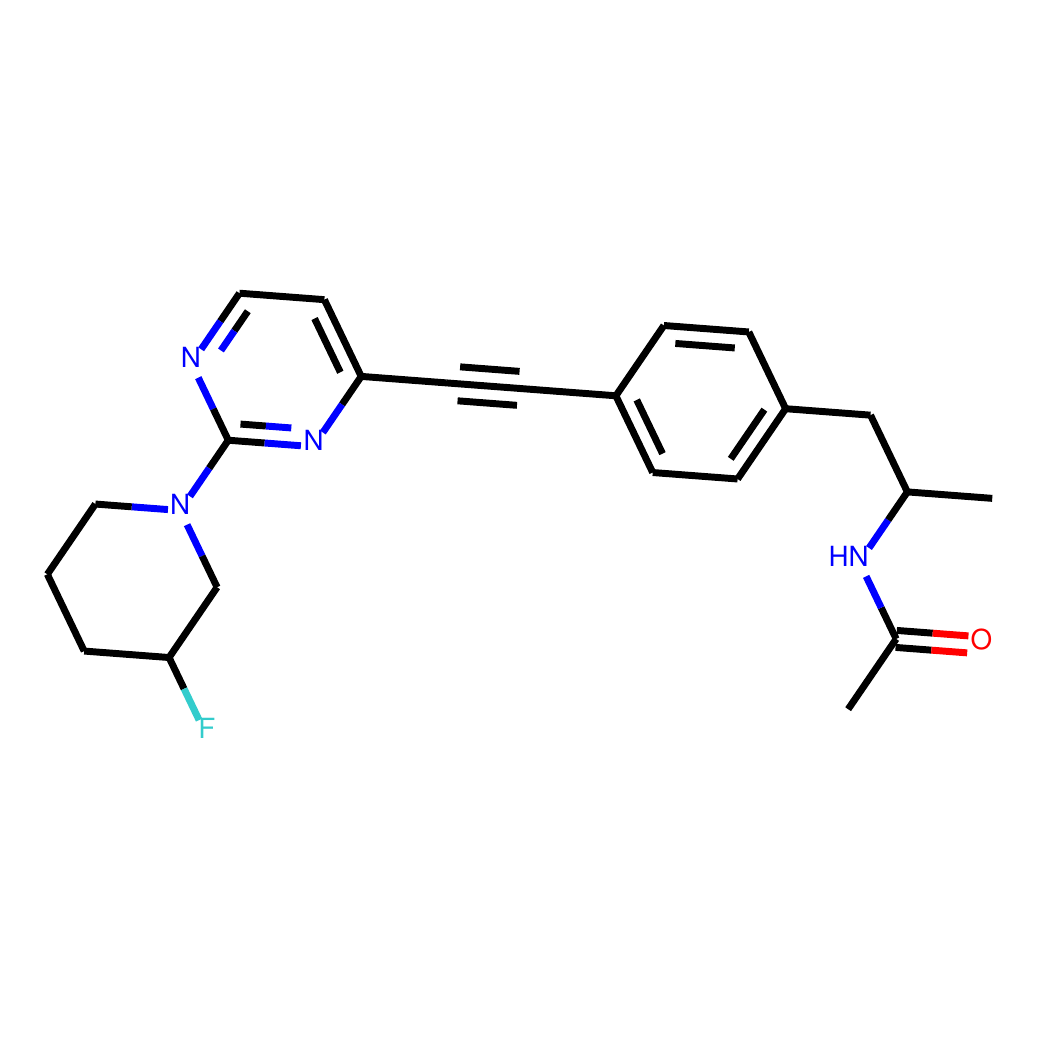} &
  \includegraphics[scale=0.18]{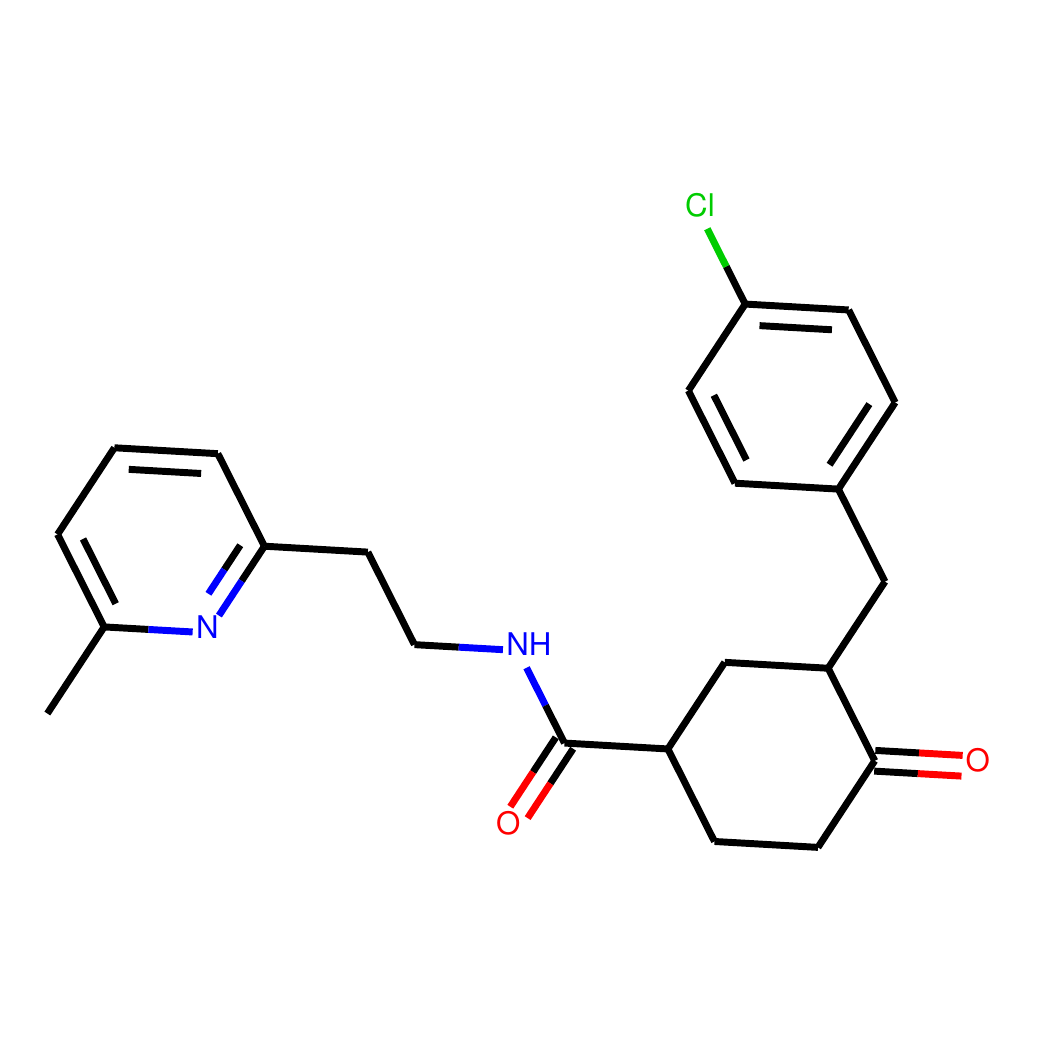} &
  \includegraphics[scale=0.18]{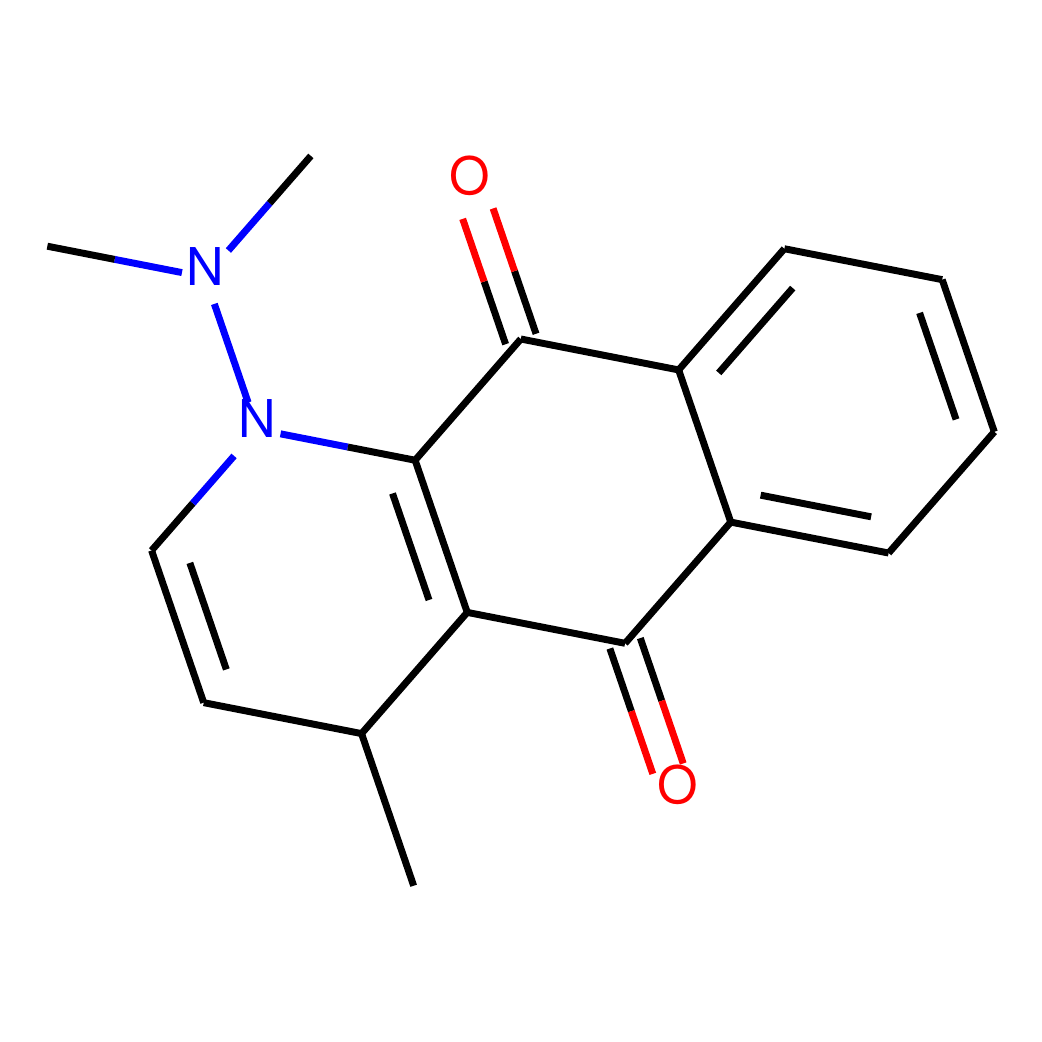} &
  \includegraphics[scale=0.18]{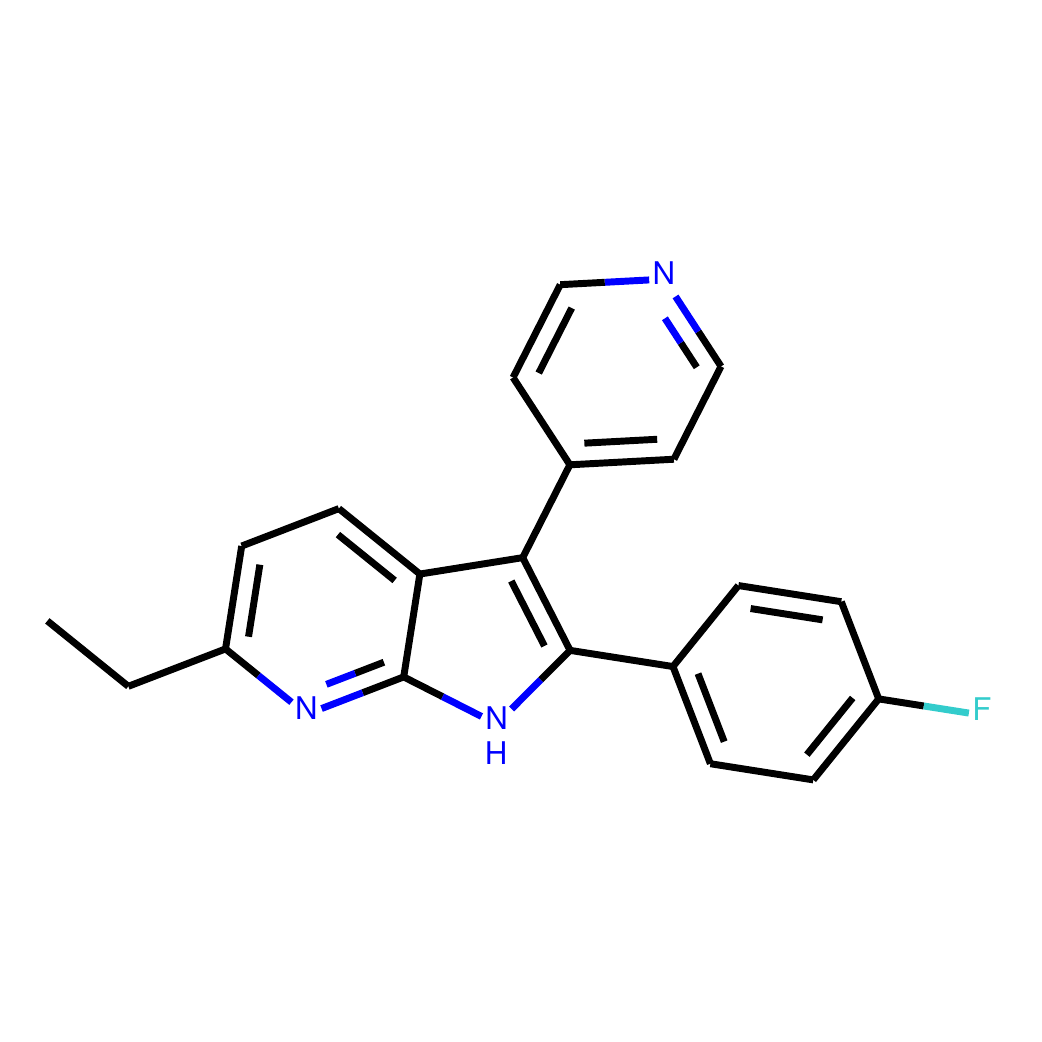} &
  \includegraphics[scale=0.18]{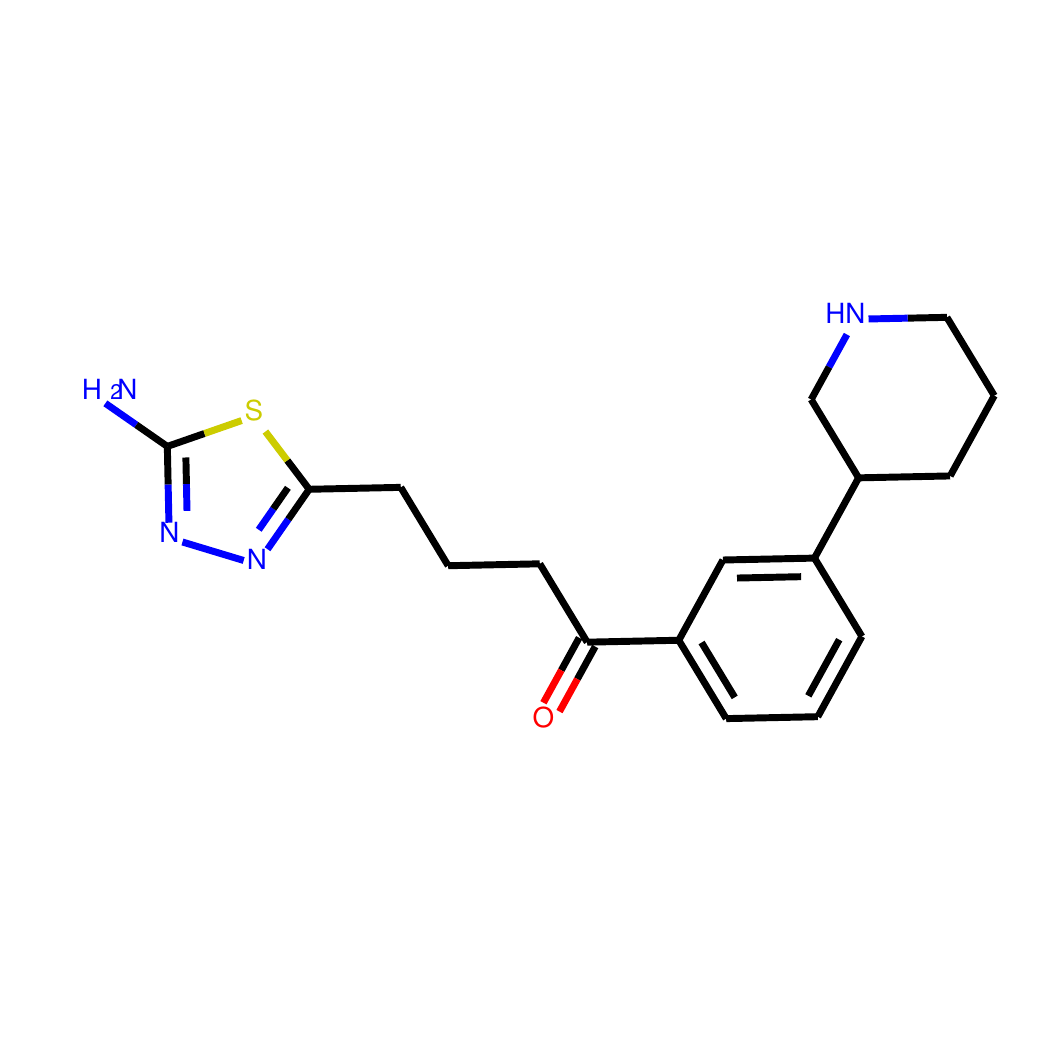} \\
  \includegraphics[scale=0.18]{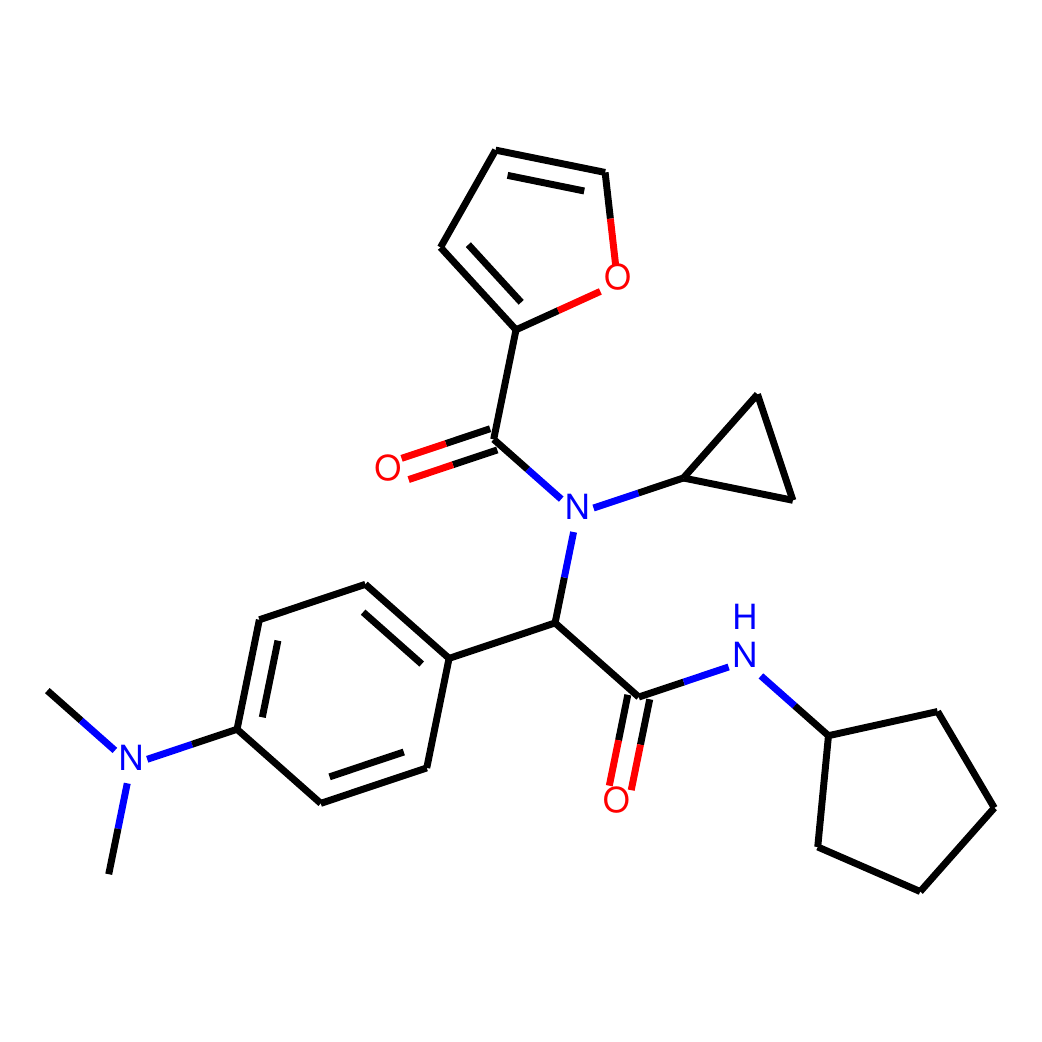} &
  \includegraphics[scale=0.18]{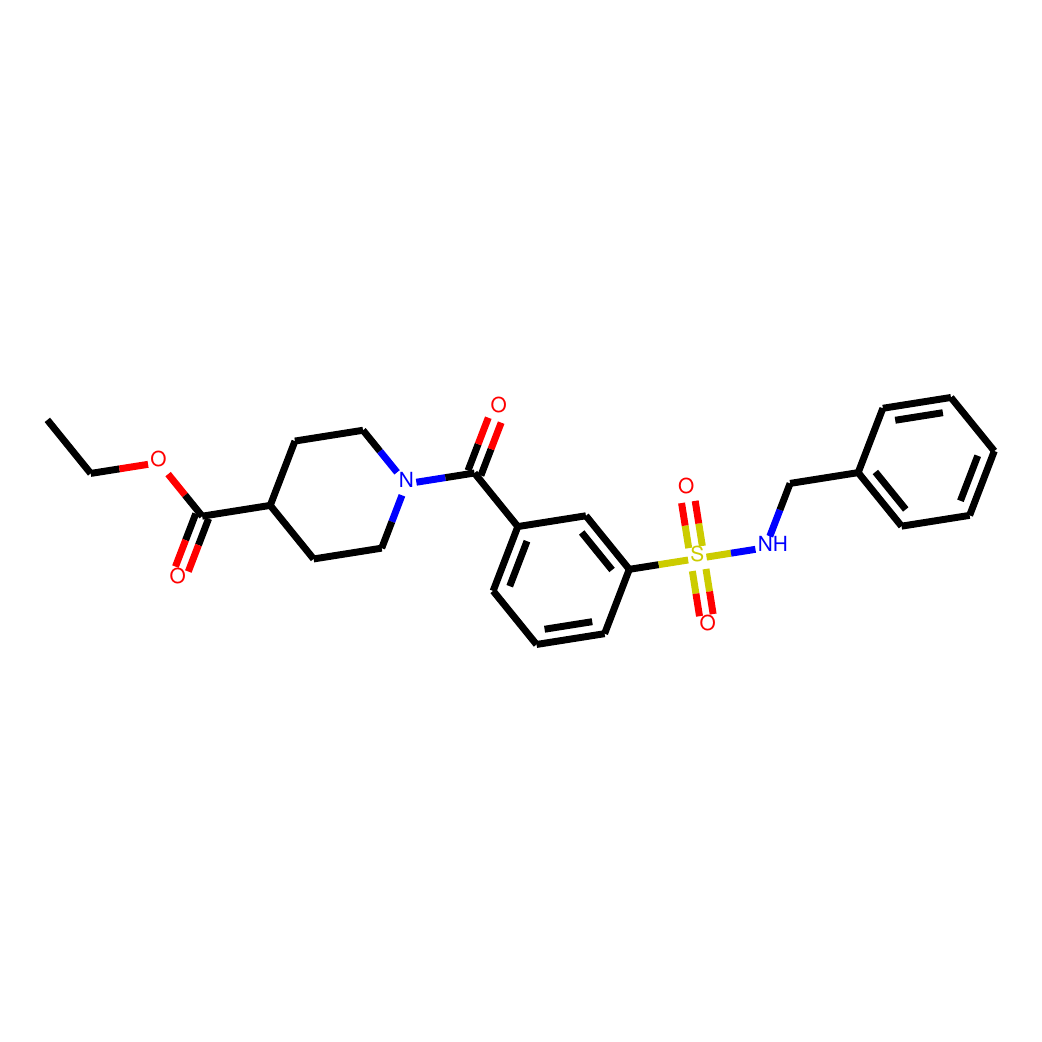} &
  \includegraphics[scale=0.18]{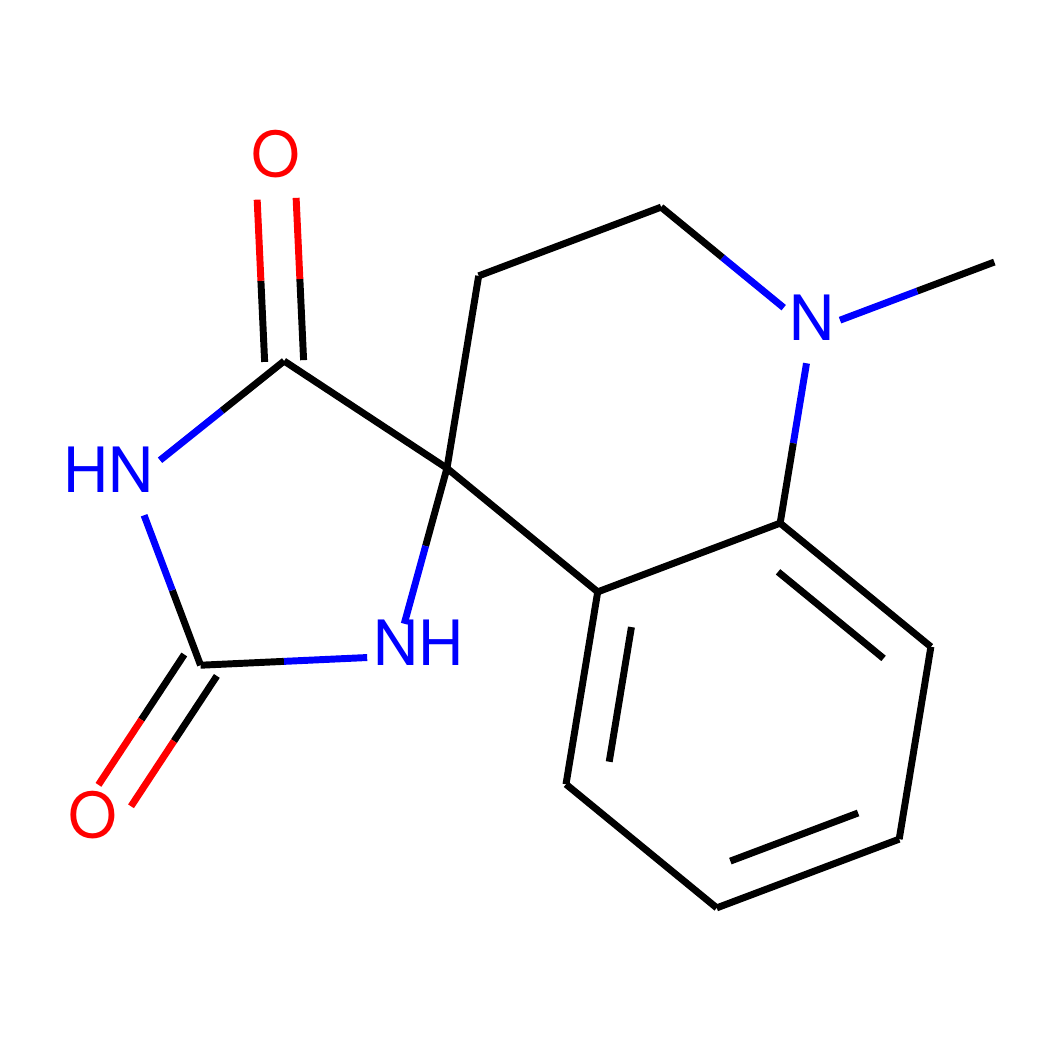} &
  \includegraphics[scale=0.18]{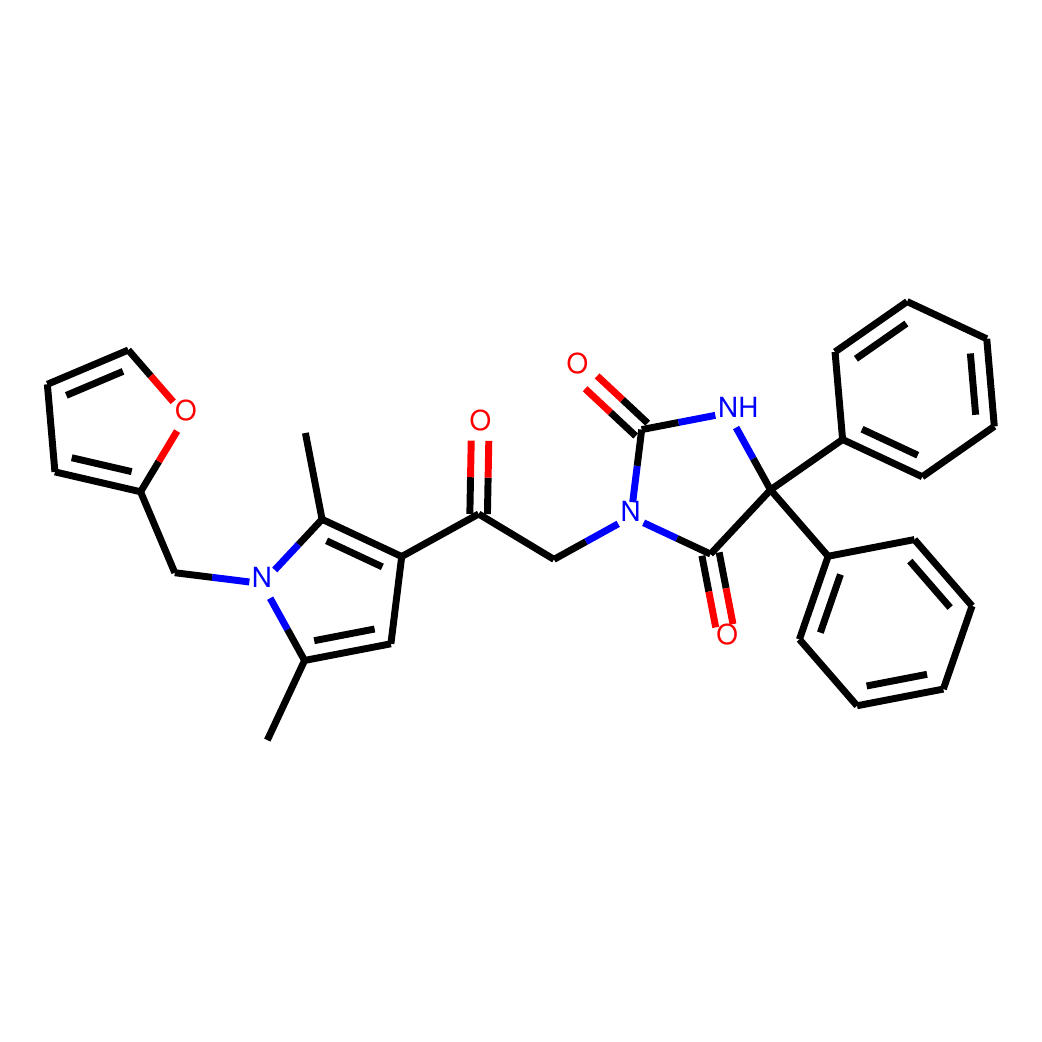} &
  \includegraphics[scale=0.18]{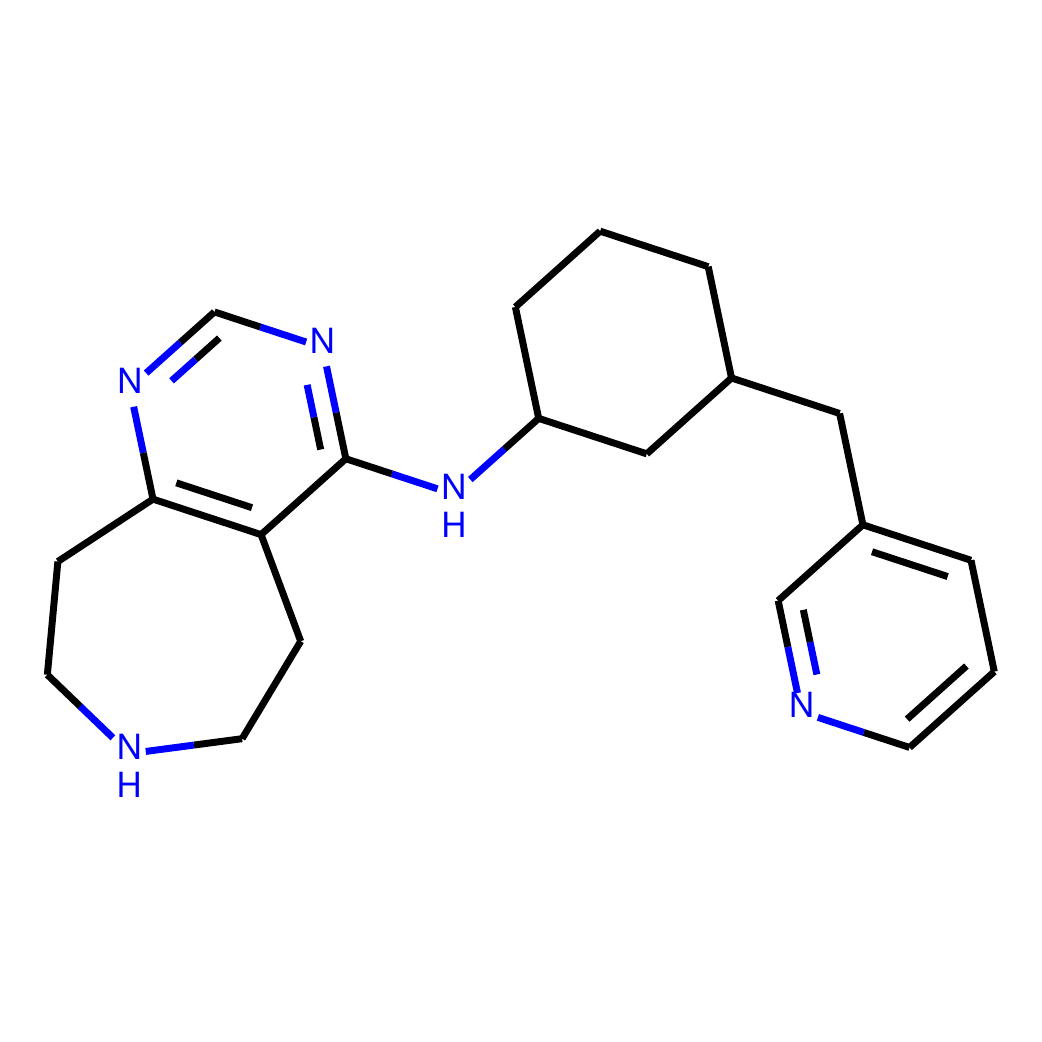} &
  \includegraphics[scale=0.18]{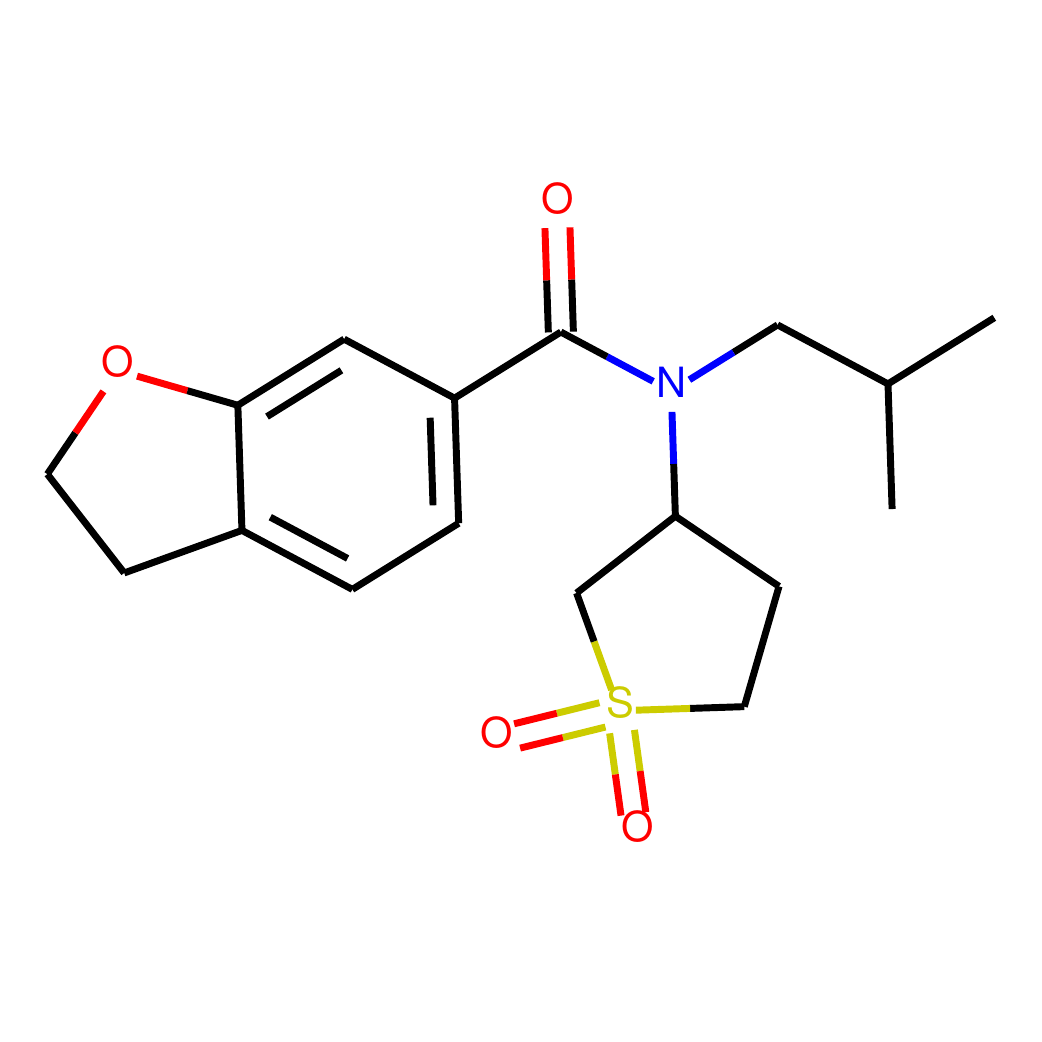} \\
  \includegraphics[scale=0.18]{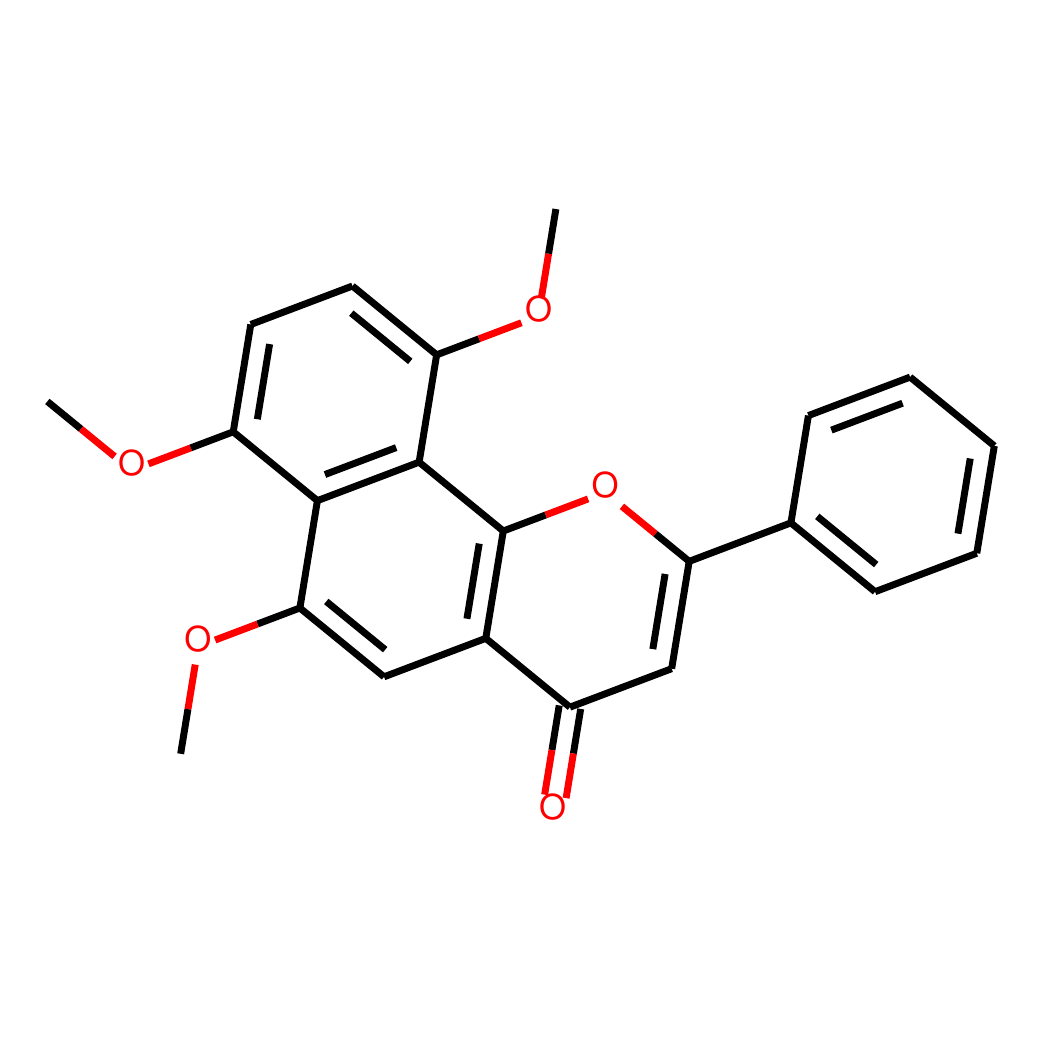} &
  \includegraphics[scale=0.18]{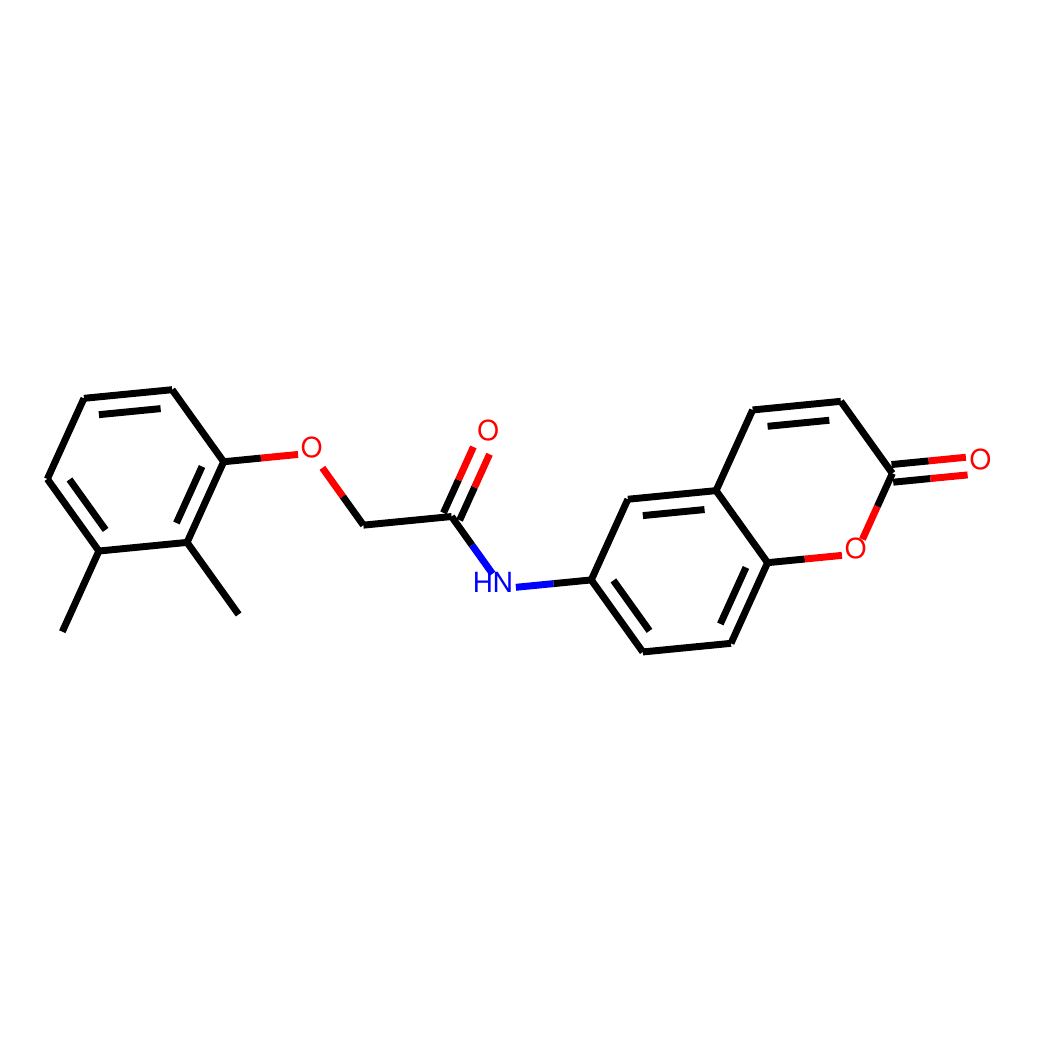} &
  \includegraphics[scale=0.18]{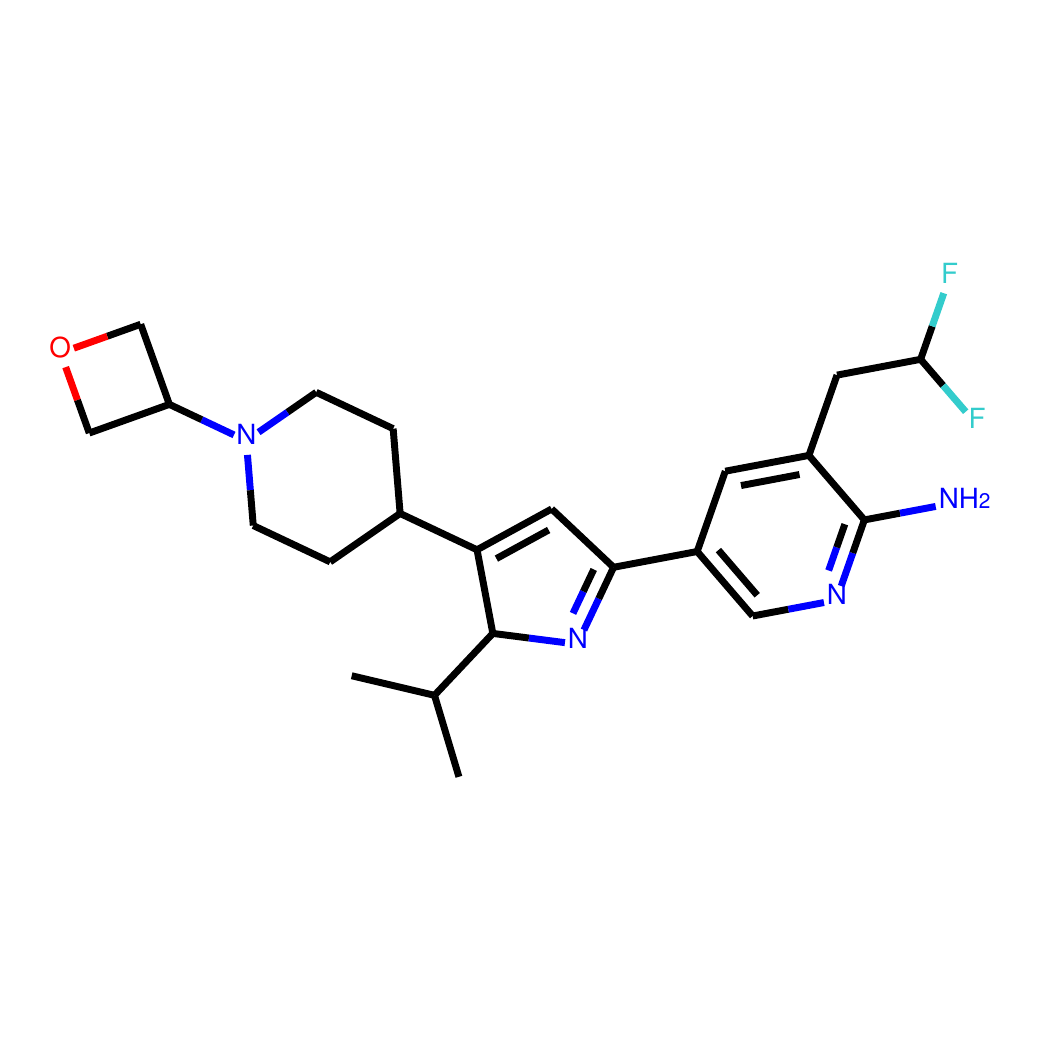} &
  \includegraphics[scale=0.18]{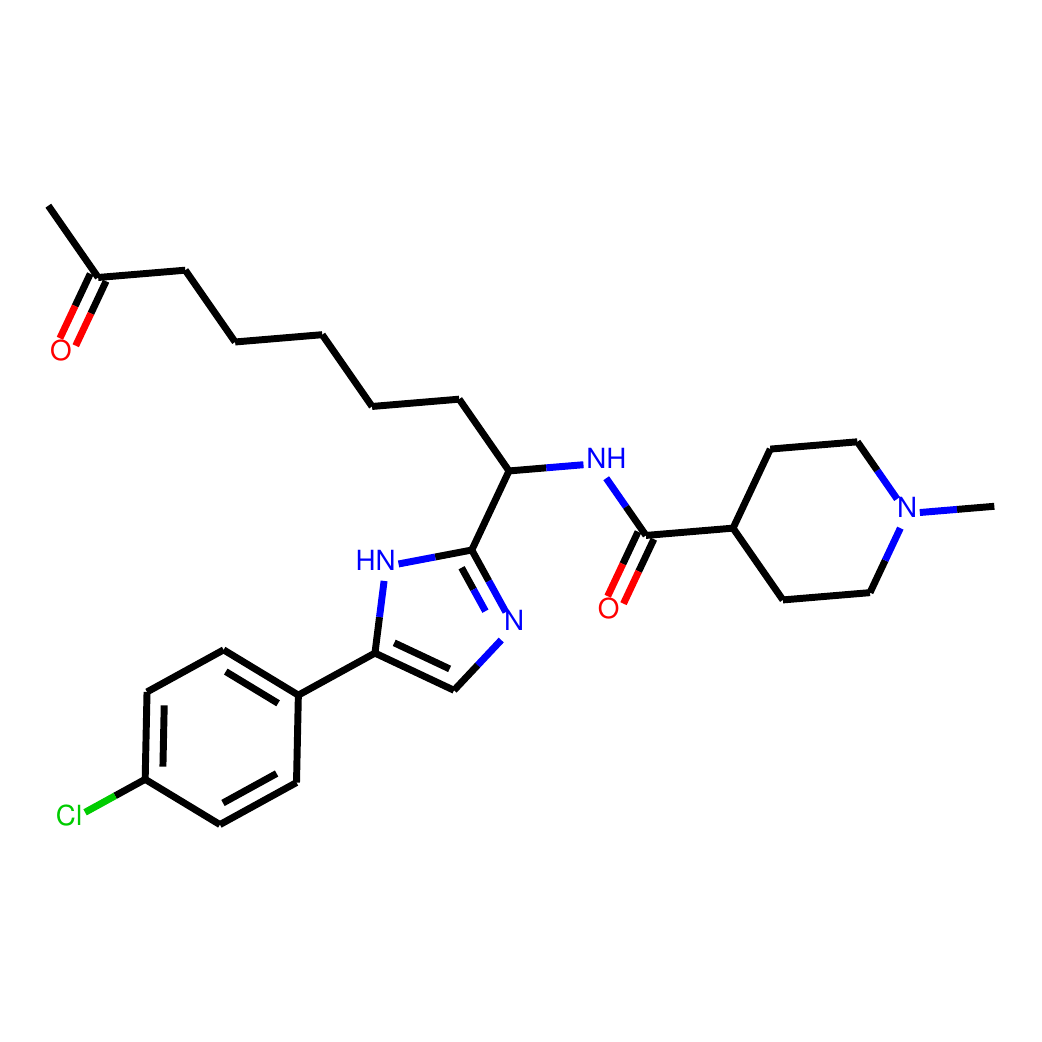} &
  \includegraphics[scale=0.18]{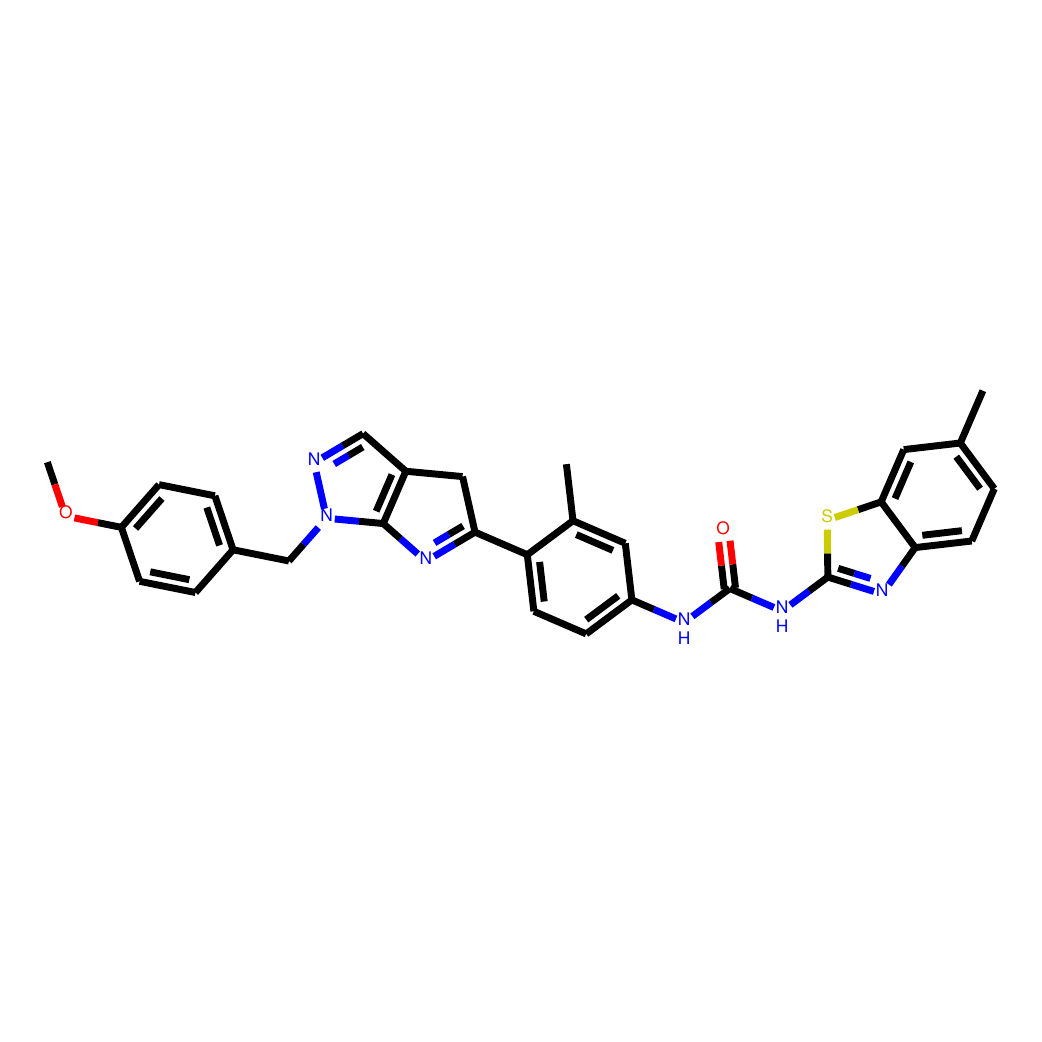} &
  \includegraphics[scale=0.18]{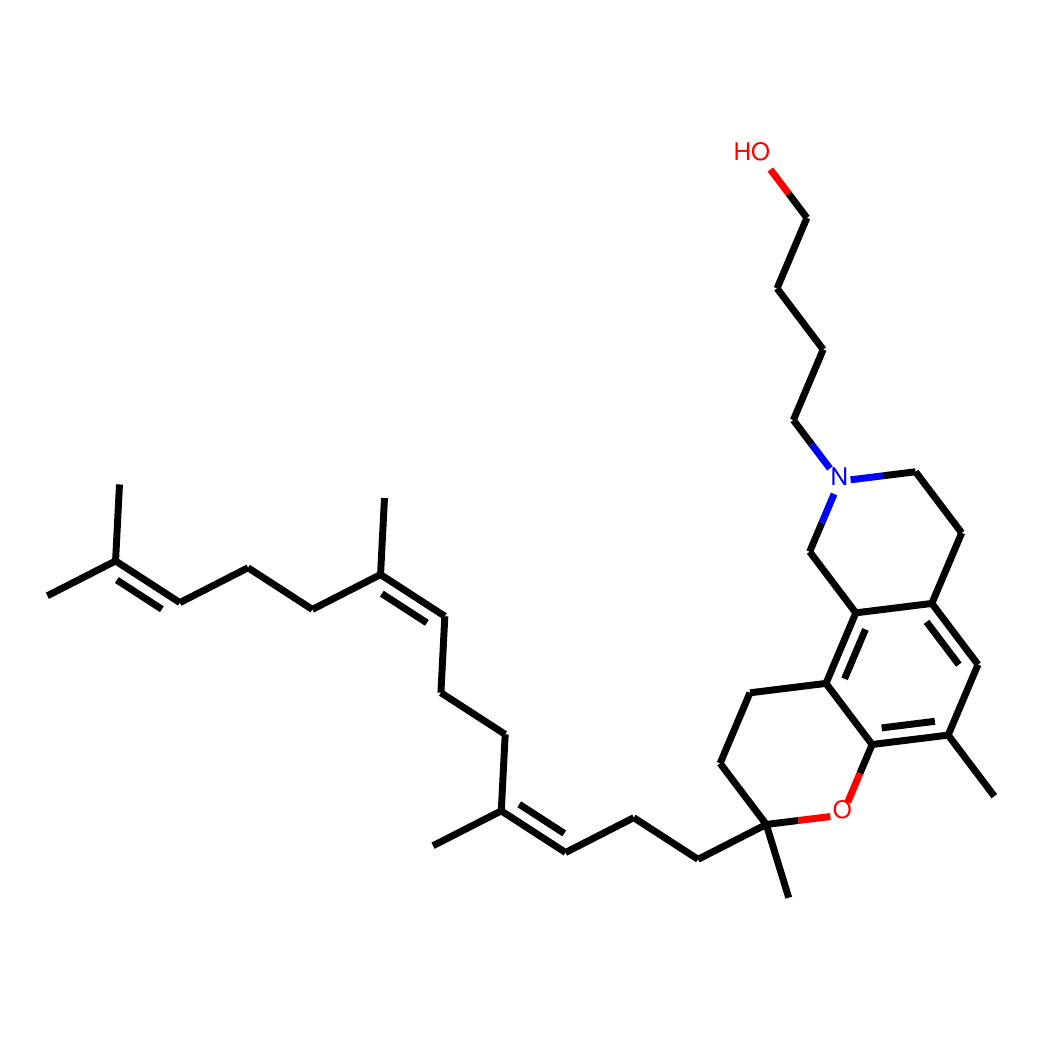} \\
  \includegraphics[scale=0.18]{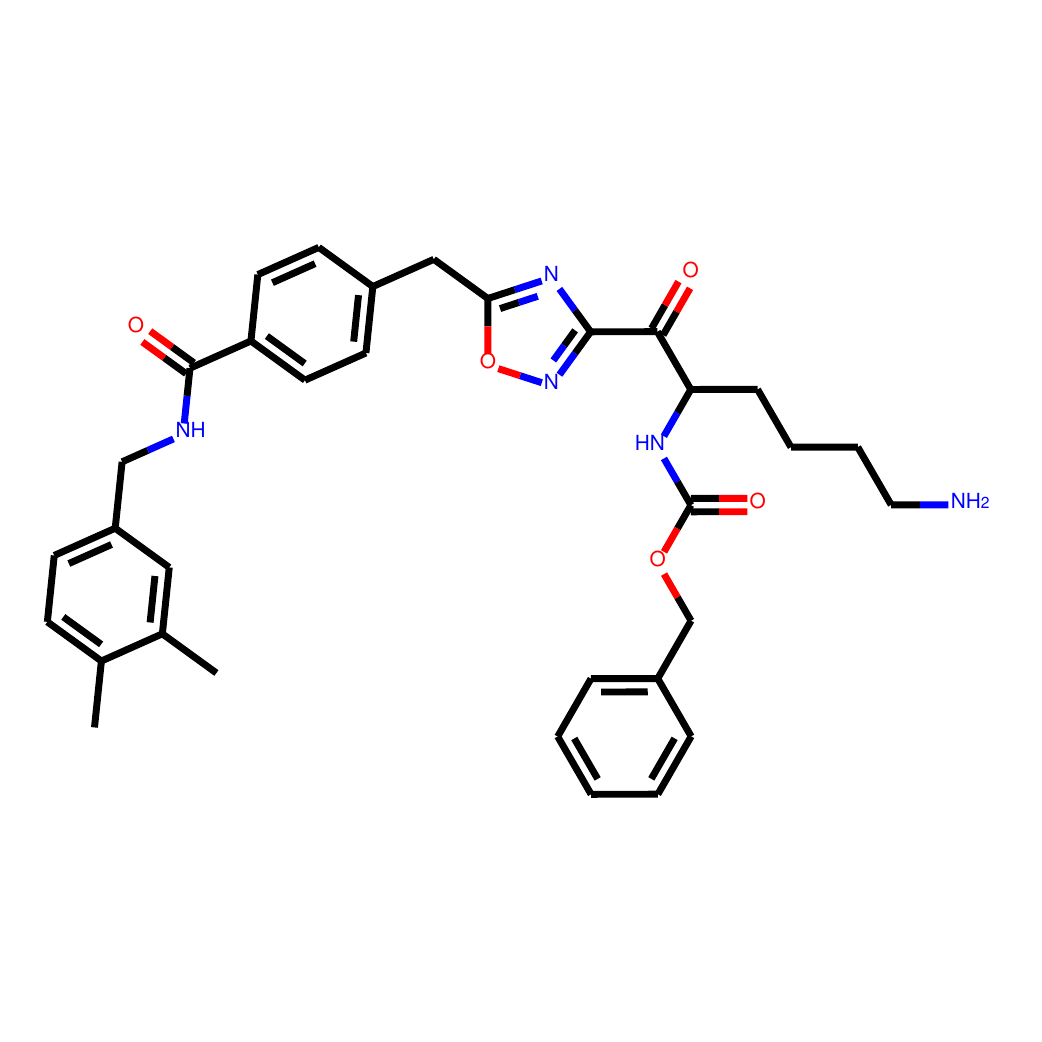} &
  \includegraphics[scale=0.18]{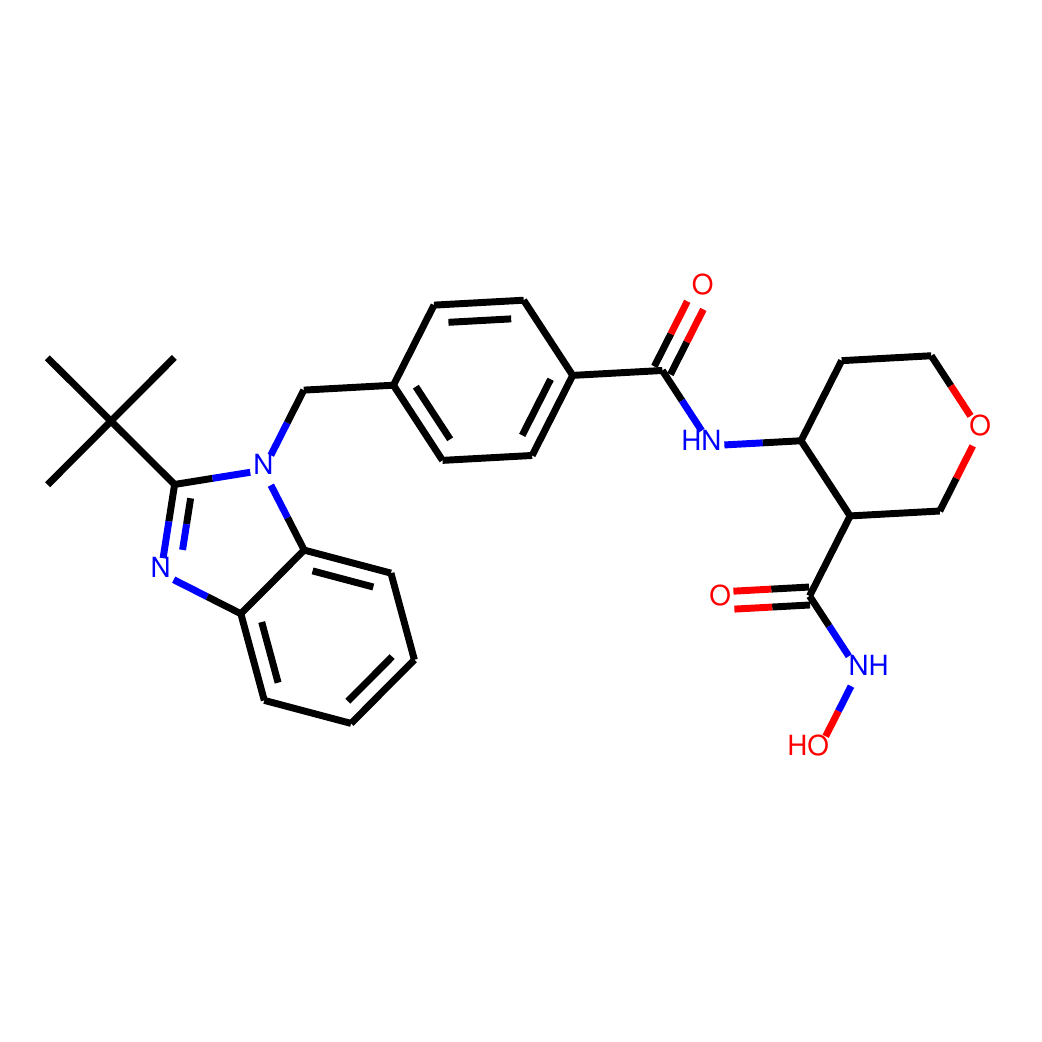} &
  \includegraphics[scale=0.18]{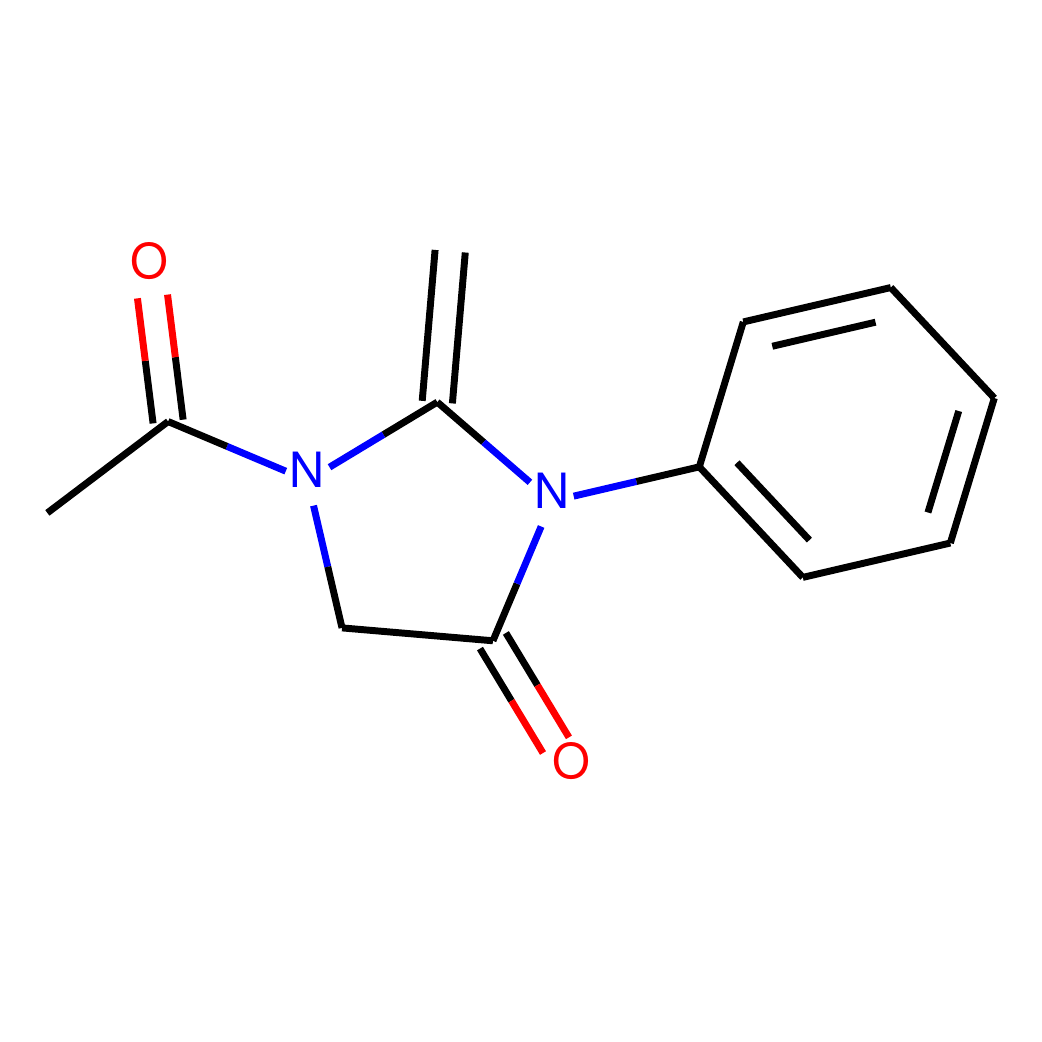} &
  \includegraphics[scale=0.18]{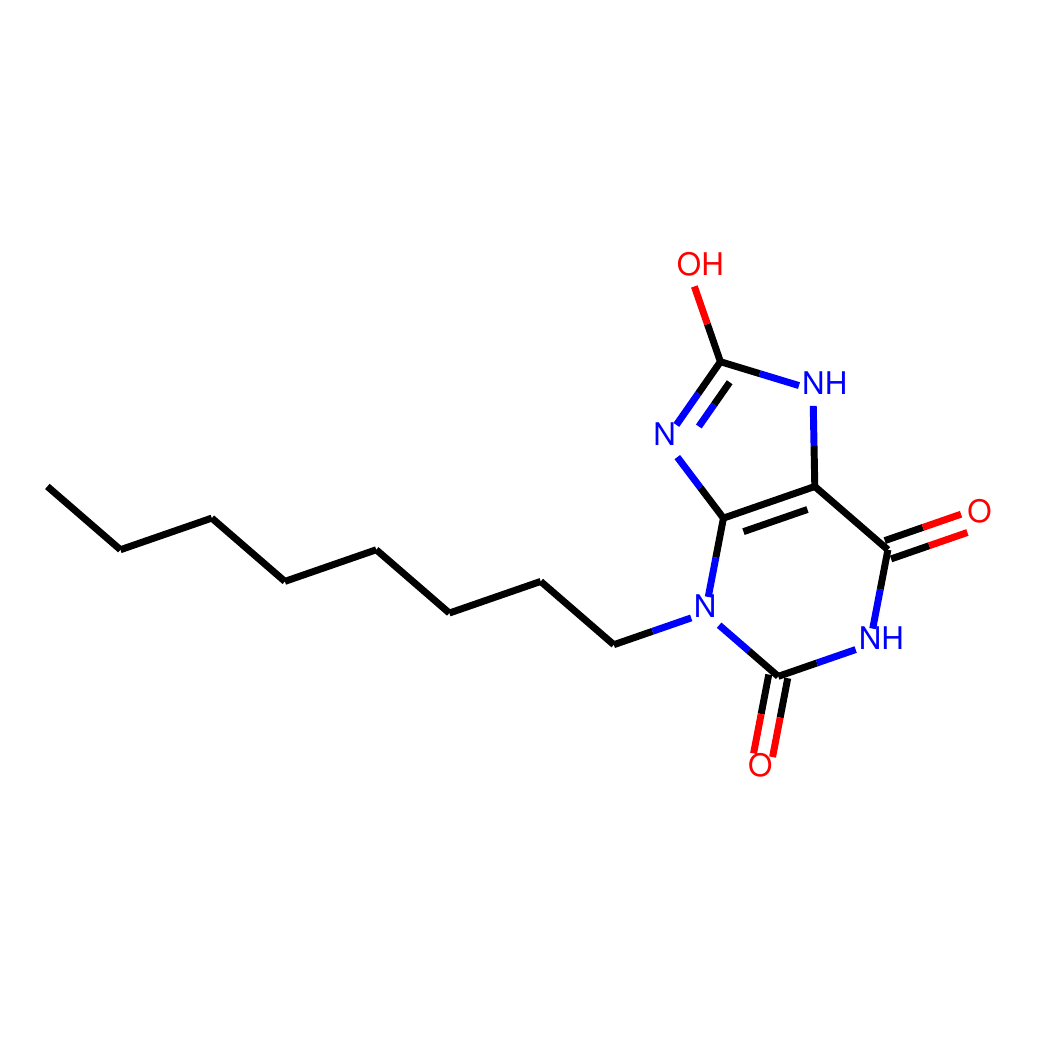} &
  \includegraphics[scale=0.18]{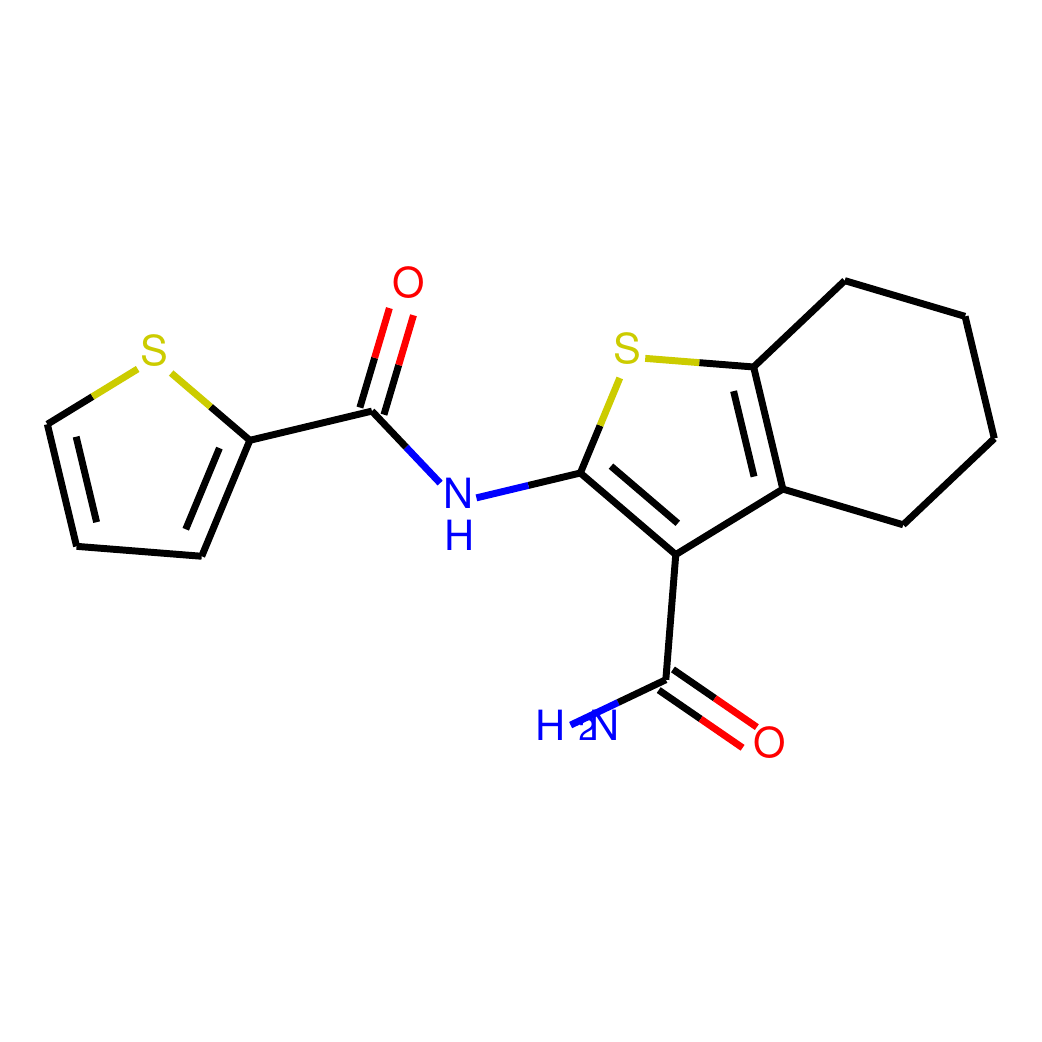} &
  \includegraphics[scale=0.18]{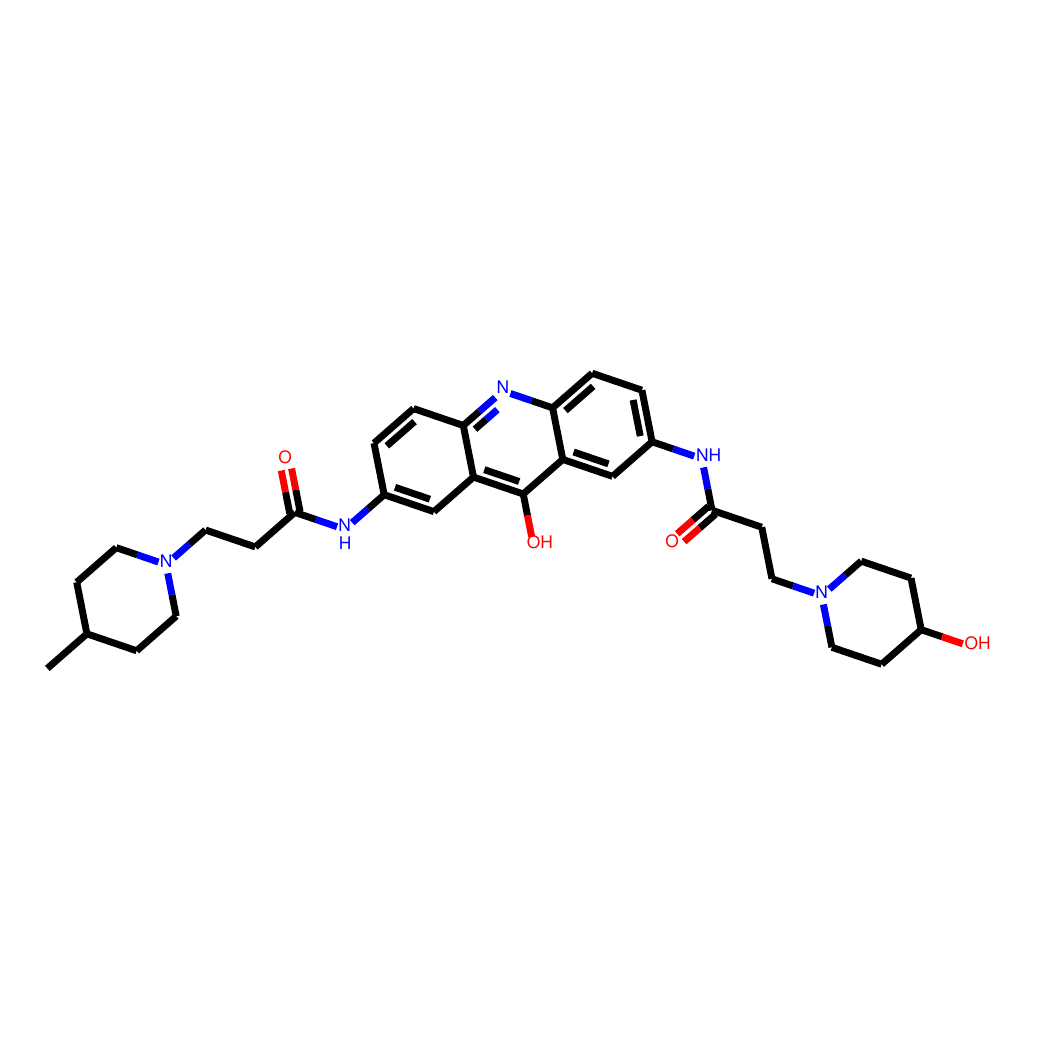} \\
  \includegraphics[scale=0.18]{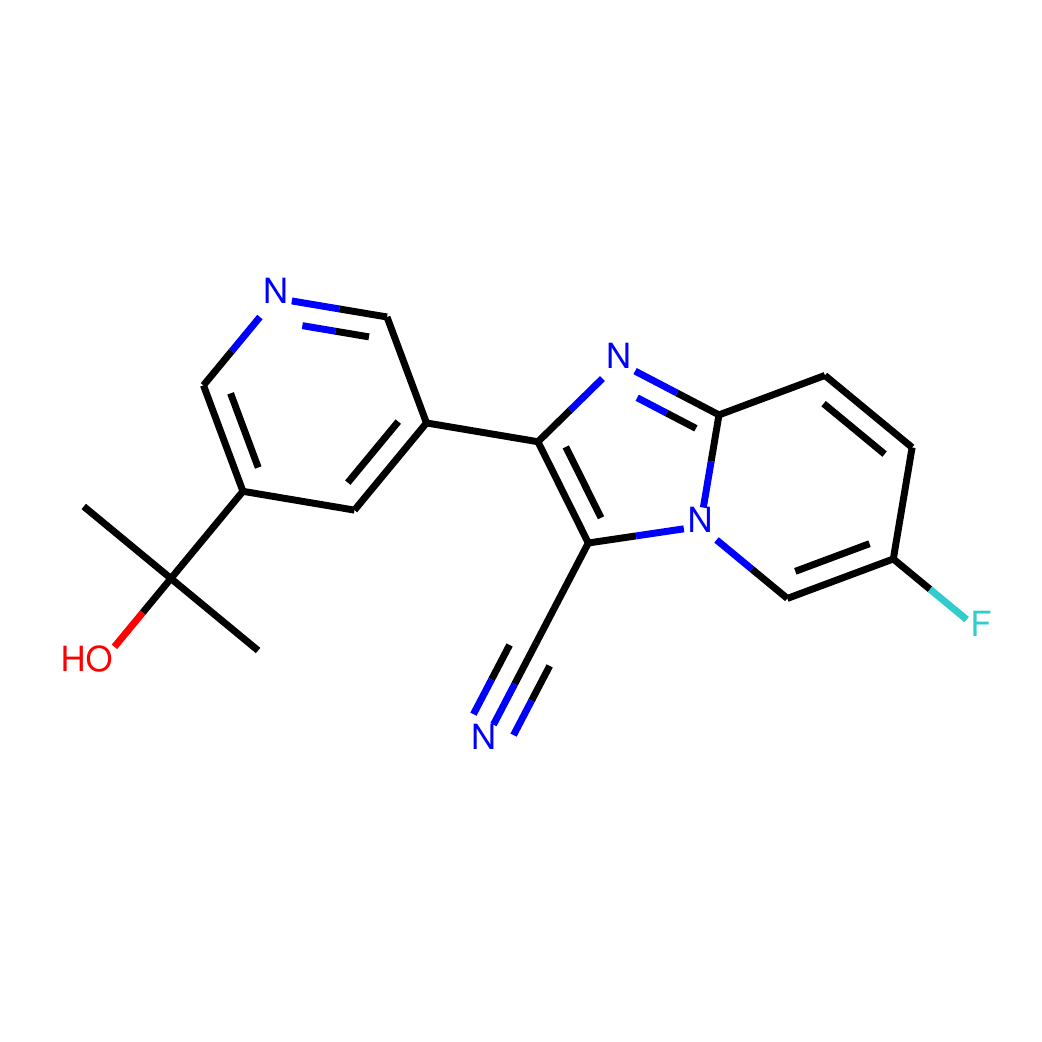} &
  \includegraphics[scale=0.18]{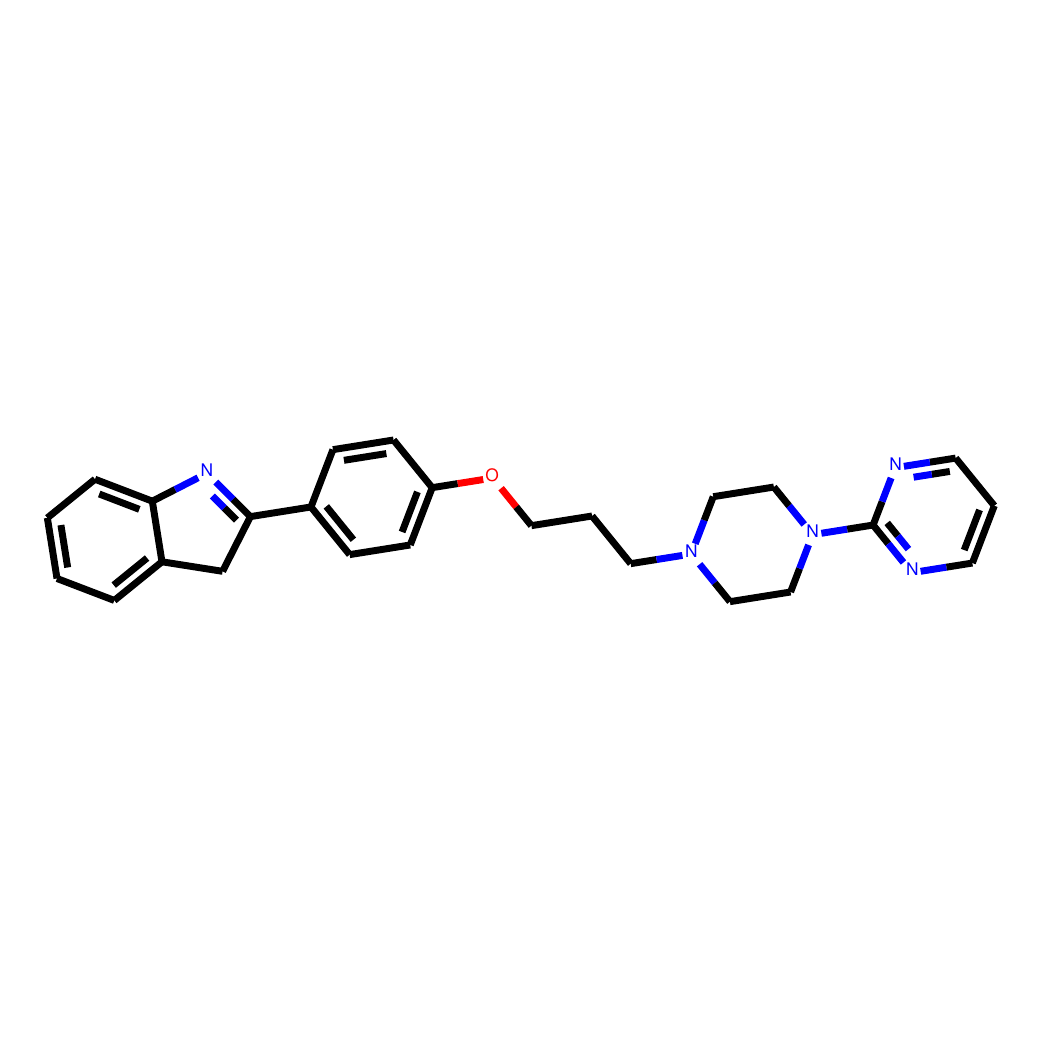} &
  \includegraphics[scale=0.18]{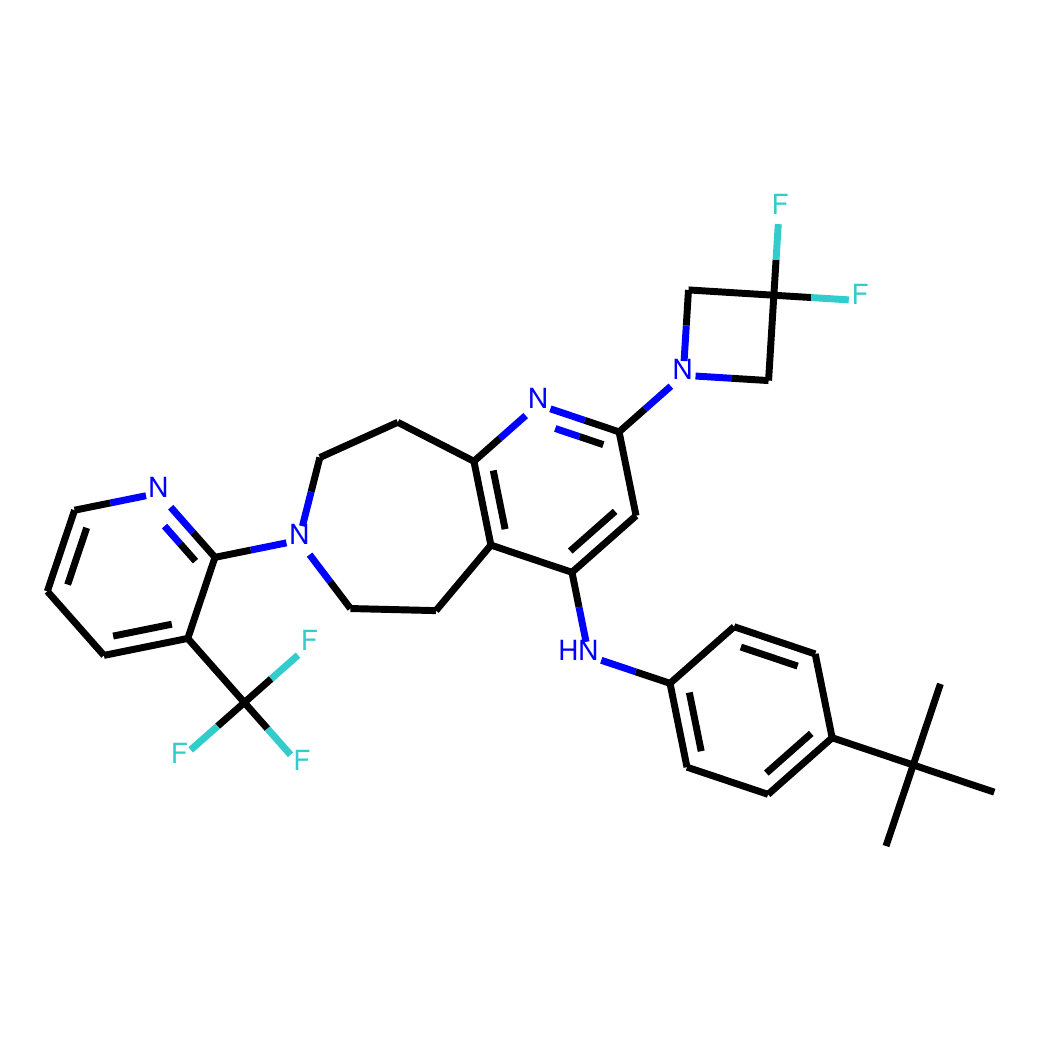} &
  \includegraphics[scale=0.18]{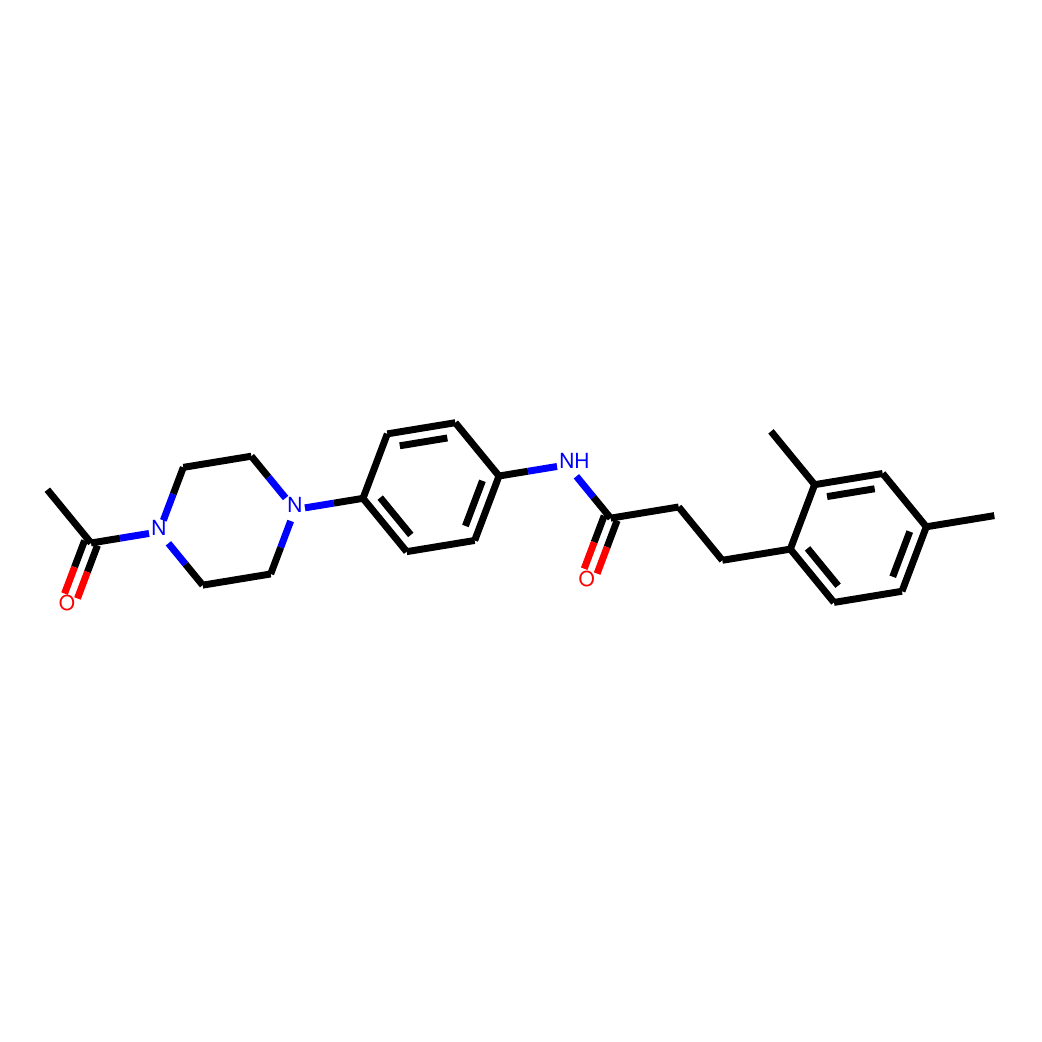} &
  \includegraphics[scale=0.18]{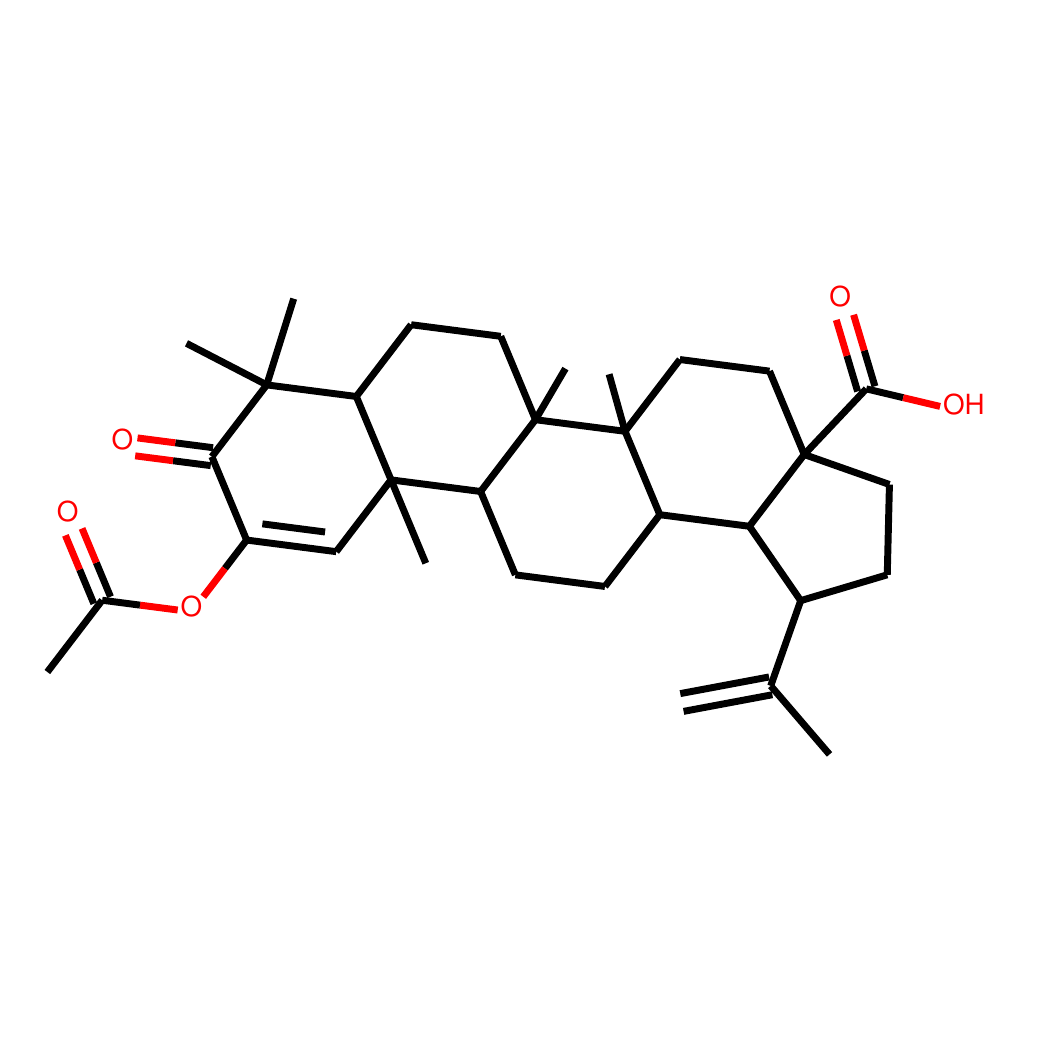} &
  \includegraphics[scale=0.18]{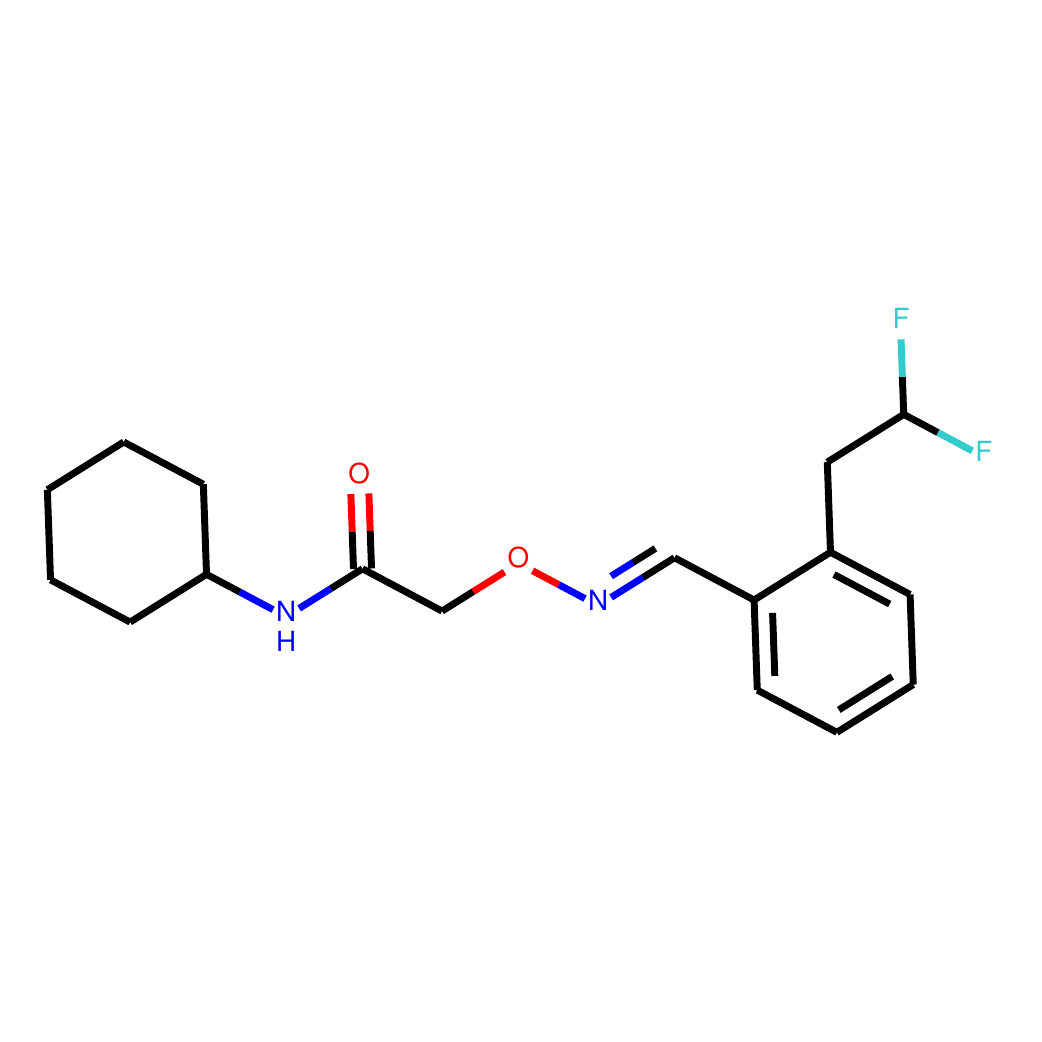} \\
  \includegraphics[scale=0.18]{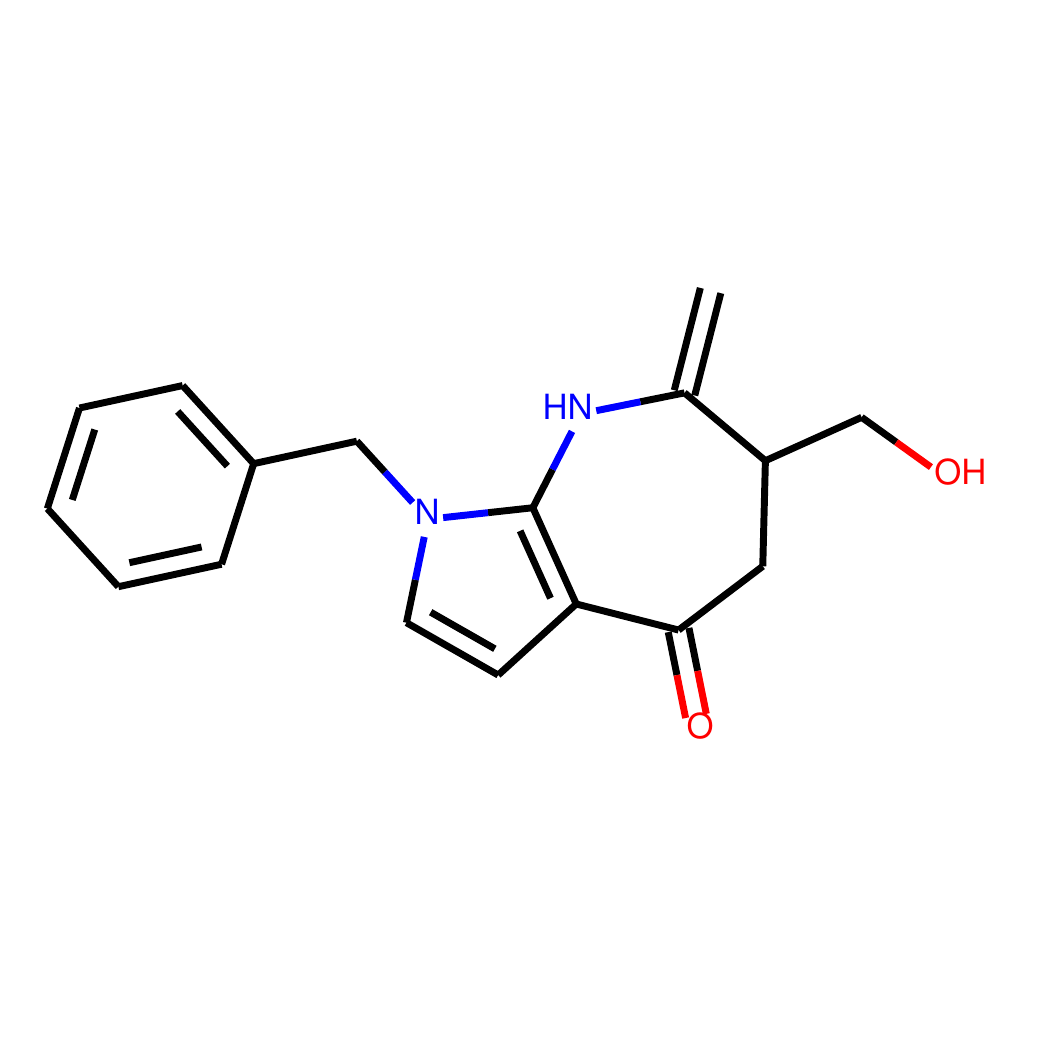} &
  \includegraphics[scale=0.18]{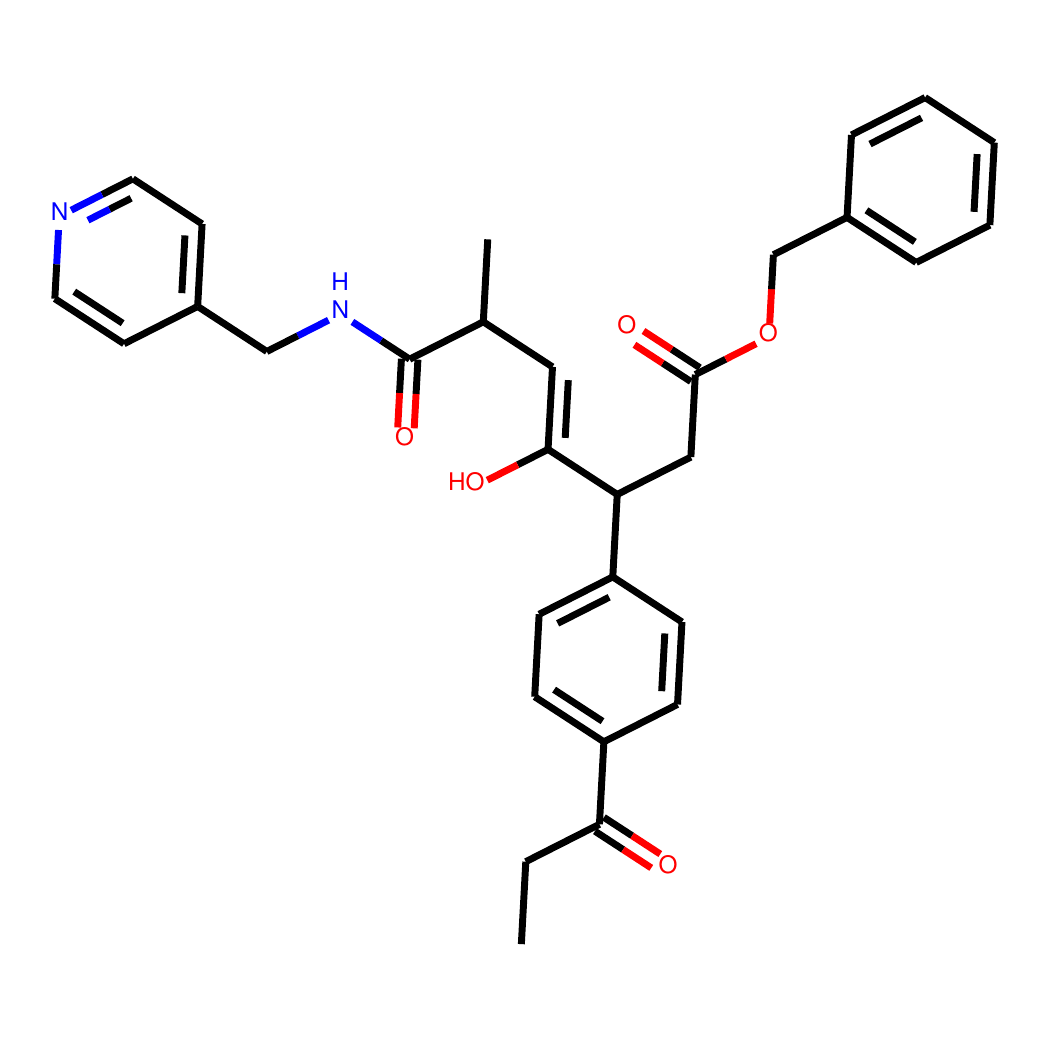} &
  \includegraphics[scale=0.18]{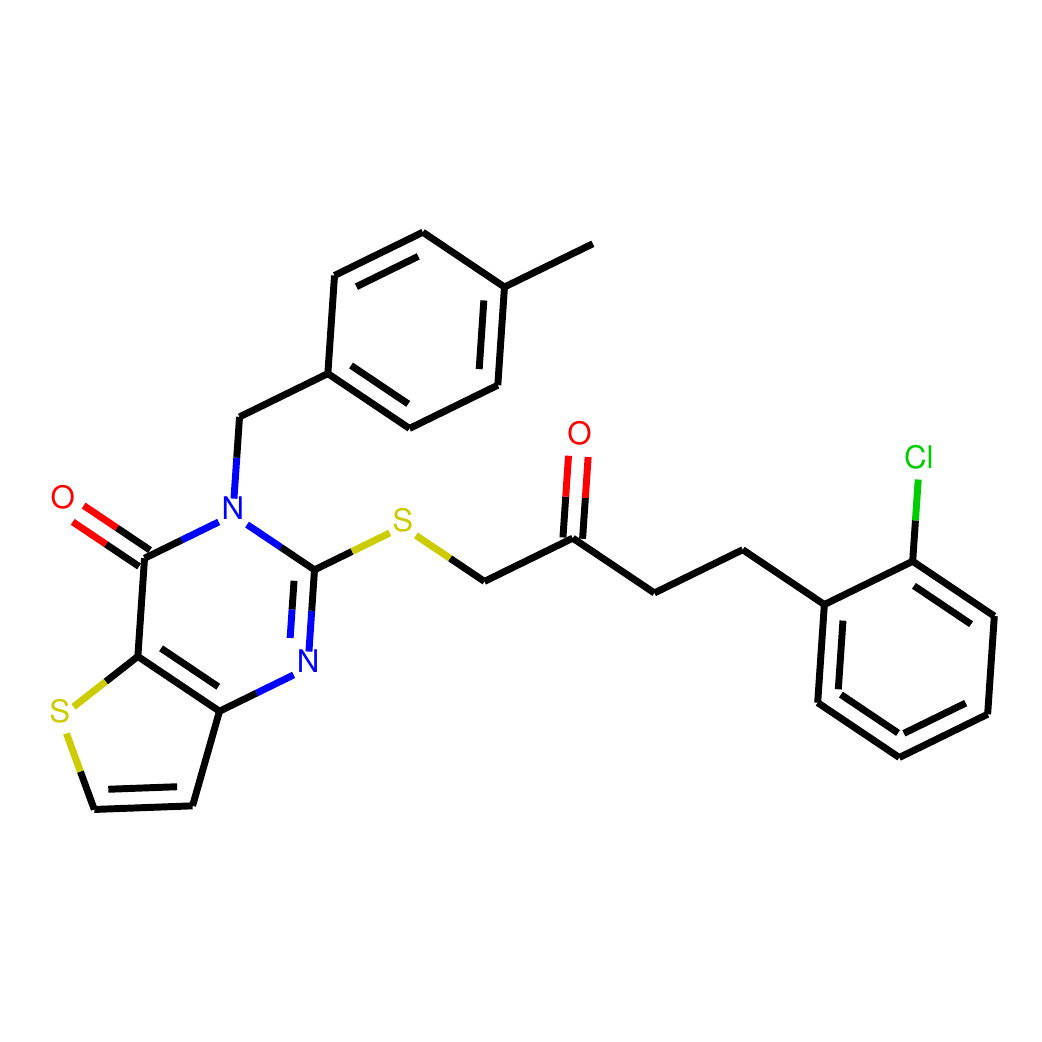} &
  \includegraphics[scale=0.18]{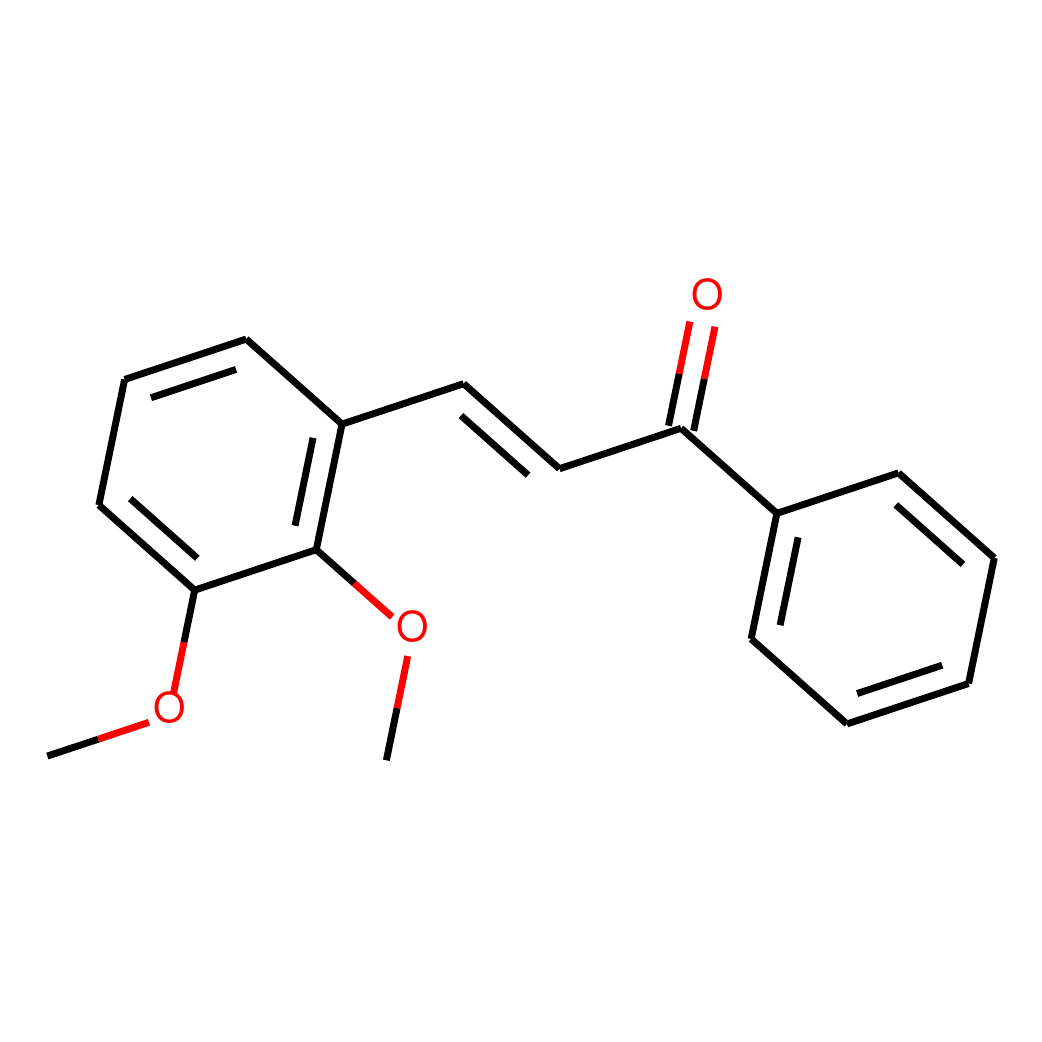} &
  \includegraphics[scale=0.18]{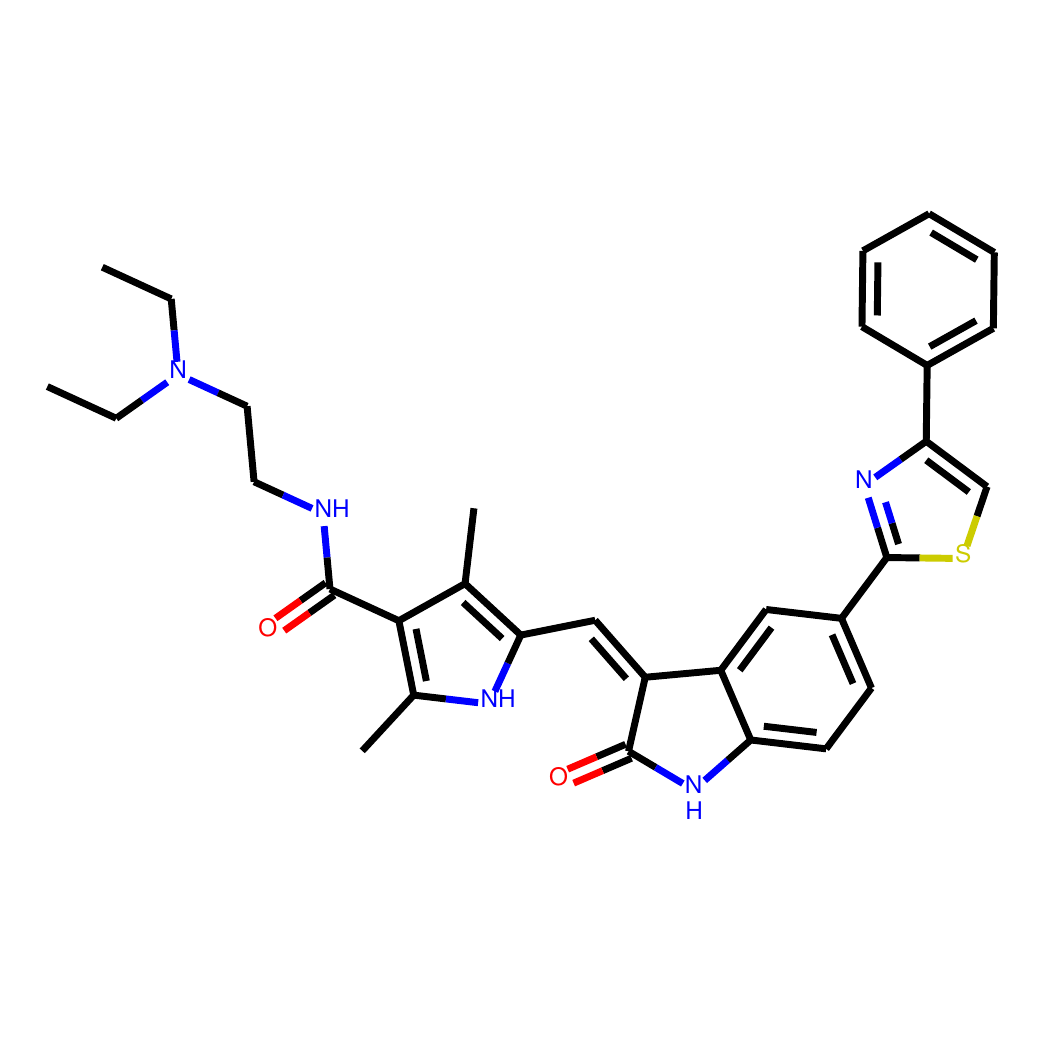} &
  \includegraphics[scale=0.18]{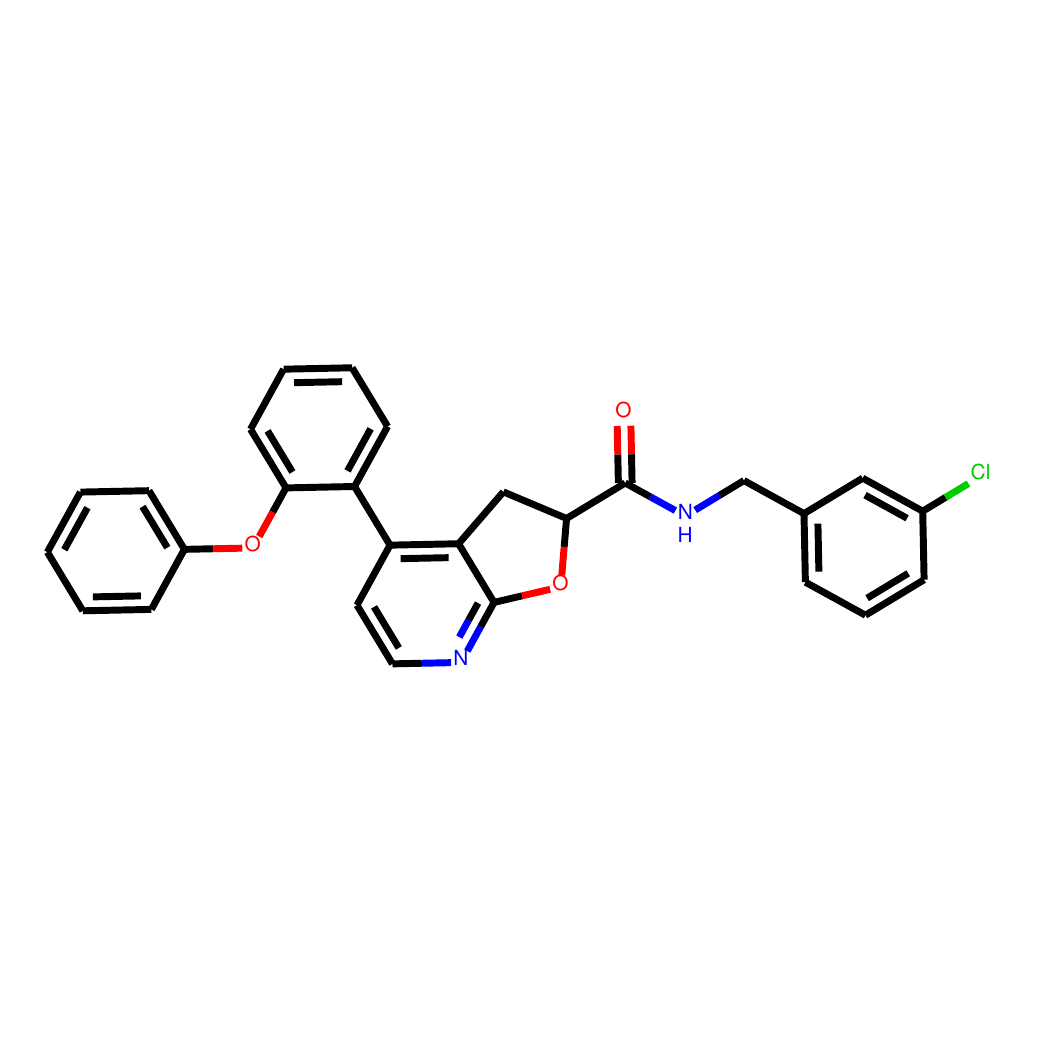} \\
\end{tabular}
\caption{Molecules generated with GCE-2 method (two-shots), model trained on ChEMBL dataset}
\end{figure}

\section{Architectures and hyperparameters for experiments with MoleculeNet}

As mentioned in the main text, the architectures and hyperparameters used for the experiments on molecule property prediction for MoleculeNet datasets follow the available implementation of HIMP. A different configuration is used per dataset:

\begin{itemize}
    \item MUV dataset: 3 layers, 300 hidden channels, dropout of p=0.185
    \item Tox21 dataset: 8 layers, 256 hidden channels, dropout of p=0.5
    \item ToxCast dataset: 3 layers, 100 hidden channels, dropout of p=0.5
    \item HIV dataset: 8 layers, 256 hidden channels, dropout of p=0.5
    \item PCBA dataset: 3 layers, 300 hidden channels, dropout of p=0.185

\end{itemize}

Each layer consists of a GINe type convolution followed by batch normalization, ReLU activation and dropout.
The final layer consists of a global mean pooling and a linear classifier.

\section{Processing time estimation for pre-training and classification}

Training time varies greatly for different graph datasets and for the different experiments performed in this work. In the main text we have discussed inference time for molecule generation. 

To give an approximate measure of processing time of the proposed method for pre-training and classification, for the dataset Graph MNIST, to perform the pretraining of the inpainting model for 100 epochs it takes approximately 30 minutes. To train the classifier for 100 epochs also takes approximately 30 minutes, taking in total approximately 1 hour for the total pretraining and classification. The hardware used for training and inference is a NVIDIA GeForce RTX 2080 Ti.

\end{document}